\documentclass[11pt,a4paper]{article}
\usepackage[utf8]{inputenc}
\usepackage[T1]{fontenc}
\usepackage[margin=0.8in]{geometry}
\usepackage{amsmath,amssymb}
\usepackage{microtype}
\usepackage{booktabs}
\usepackage{graphicx}
\usepackage[safe]{tipa}
\usepackage{algorithm}
\usepackage{algpseudocode}
\usepackage[round,authoryear]{natbib}
\usepackage[colorlinks=true,linkcolor=blue,citecolor=blue,urlcolor=blue]{hyperref}

\title{LipSpell: A Language-Independent Phonetic Scoring Framework for Spelling Correction Combining Distinctive-Feature Edit Distance, Keyboard Geometry, and Corpus Frequency}
\author{
  Leonardo Araujo\\
  Department of Computer Science, Universidade Federal de S\~ao Jo\~ao del-Rei (UFSJ)\\
  S\~ao Jo\~ao del-Rei, MG, Brazil\\
  \texttt{leolca@ufsj.edu.br}
}
\date{November 2026}

\begin{document}

\maketitle

\begin{abstract}
Automatic spelling correction must handle both typographical slips on a keyboard and cognitive or phonetic misspellings in which writers approximate a word's pronunciation. Conventional spell checkers rely either on orthographic edit distance and keyboard proximity---which struggle when phonetic errors diverge by multiple characters---or on English-centric phonetic hashes such as Soundex and Metaphone. We present \textbf{LipSpell}, a unified, language-independent phonetic scoring engine paired with swappable language-specific G2P frontends, keyboard maps, and lexicons that integrates: (1) a precomputed $|P|\times|P|$ phone-distance matrix derived from universal IPA distinctive features, (2) physical keyboard geometry, and (3) corpus unigram frequency priors, accelerated by a two-stage symmetric-delete and seven-class phonetic-code index. Across six English benchmarks (including $1{,}000$ L2 learner errors from TOEFL-Spell), LipSpell outperforms Norvig's baseline and Hunspell on every dataset and beats Aspell on Norvig1 ($78.52\%$ vs.\ $75.56\%$), Holbrook ($32.00\%$ vs.\ $26.67\%$), and TOEFL-Spell ($77.30\%$ vs.\ $72.80\%$). Evaluated across five additional languages---together with English, spanning six languages, five linguistic branches (Germanic, Romance, Slavic, Semitic, and Turkic), three top-level families (Indo-European, Afro-Asiatic, and Turkic), and three writing systems---LipSpell achieves strong fixed-weight cross-lingual transfer on Brazilian Portuguese ($73.20\%$ Top-1 on the $n=500$ Wikipedia slice without NFD tolerance and $76.20\%$ with NFD tolerance vs.\ $66.40\%$ Norvig, $64.20\%$ Hunspell, $64.00\%$ Aspell) and Russian Cyrillic ($85.67\%$ vs.\ $78.50\%$ Norvig), while per-language lexicon cleaning, weak-character indel costs, and weight tuning on $50/50$ clean in-vocabulary non-word splits yield strong gains over Norvig across four of the six languages---Portuguese ($87.17\%$ vs.\ $76.75\%$), Turkish ($81.67\%$ vs.\ $79.05\%$), Hebrew Abjad ($74.59\%$ vs.\ $62.30\%$), and French ($74.53\%$ vs.\ $67.92\%$)---with fixed English weights remaining optimal for English ($80.37\%$) and Russian ($85.67\%$).
\end{abstract}

\section{Introduction}
Automatic spelling correction is a foundational component of text editors, search engines, optical character recognition (OCR) post-processing, educational and assistive writing tools, and noisy-text normalization pipelines \citep{kukich1992techniques}. Human non-word spelling errors fall broadly into two distinct generative mechanisms \citep{damerau1964technique,toutanova2002pronunciation}: \emph{typographical errors}, which arise from motor slips on a physical or virtual keyboard (such as hitting an adjacent key, omitting a keystroke, or transposing two consecutive characters), and \emph{cognitive or phonetic errors}, in which the writer knows or approximates the spoken form of a word but maps its phonemes onto an implausible or non-standard orthography (e.g., \textit{nite} $\rightarrow$ \textit{night}, \textit{semetary} $\rightarrow$ \textit{cemetery}, \textit{proplen} $\rightarrow$ \textit{problem}, or in Brazilian Portuguese \textit{escess\~ao} $\rightarrow$ \textit{exce\c{c}\~ao} and \textit{aterisagem} $\rightarrow$ \textit{aterrissagem}). Second-language (L2) learners, dyslexic writers, and users typing on mobile devices produce high proportions of cognitive-phonetic errors that frequently diverge from the target word by two or more grapheme edits.

Conventional spell checkers struggle to bridge both error types in a language-independent manner. Purely orthographic algorithms based on Levenshtein or Damerau--Levenshtein edit distance \citep{damerau1964technique,norvig2007spell} and symmetric-delete indexing \citep{garbe2012symspell} excel at single-edit typographical slips ($d_{\text{orth}} = 1$), yet fail sharply when a phonetic misspelling differs by multiple characters or when an intervening high-frequency orthographic competitor lies at the same or smaller character-edit distance (for example, \textit{albun} $\rightarrow$ \textit{algum} vs.\ \textit{\'album}, both at $d_{\text{orth}}=2$). Conversely, classical phonetic hashing schemes such as Soundex \citep{russell1918soundex} and Metaphone \citep{philips1990metaphone} collapse entire consonant classes into coarse English-centric codes, losing fine-grained articulatory distinctions and requiring manual redesign for every new language or script.

Our contribution lies in the principled \emph{integration and engineering} of complementary orthographic, motor-geometric, phonological, and lexical signals into \textbf{LipSpell}, a language-independent phonetic scoring framework paired with swappable, lightweight language-specific frontends and lexicons. The name \emph{LipSpell} (\textbf{L}anguage-\textbf{I}ndependent \textbf{P}honetic \textbf{Spell}ing) evokes both the articulatory organs of speech production ($[+\text{labial}]$) and a phonetic misspelling pun on \emph{leap} (e.g., \textit{lip ov fayth} $\rightarrow$ \textit{leap of faith}), where /\textsci/ $\leftrightarrow$ /i/ and /v/ $\leftrightarrow$ /f/ differ by only a single distinctive feature (tense/lax and voicing, respectively). Rather than treating phoneme substitutions as binary equal/unequal events, LipSpell precomputes a $|P|\times|P|$ phone-distance matrix directly from universal International Phonetic Alphabet (IPA) distinctive features and embeds it inside a generalized weighted Damerau--Levenshtein dynamic program. To scale to lexicons of $75\text{k}$--$777\text{k}$ words without scanning the entire vocabulary, LipSpell pairs a $0$--$2$-character symmetric-delete index with a seven-class phonetic-code index derived directly from the distinctive-feature table.

Specifically, we investigate three research questions:
\begin{enumerate}
    \item[\textbf{RQ1:}] Does a universal distinctive-feature phone representation---and which feature-vector distance metric ($L_1$, salience-weighted $L_1$, or cosine)---maximize phonetic spelling correction and two-stage candidate recall?
    \item[\textbf{RQ2:}] Are orthographic edit distance, physical keyboard geometry (with diacritic tolerance), distinctive-feature phonetic distance, and corpus frequency priors complementary when combined in a unified scoring function?
    \item[\textbf{RQ3:}] Can the same phonetic scoring and indexing engine transfer across multiple typologically diverse languages, language families, and writing systems when only the G2P frontend, keyboard layout, and lexicon are replaced?
\end{enumerate}

The remainder of this article is organized as follows. Section~\ref{sec:related} situates LipSpell within prior work on noisy-channel spelling correction, phonetic hashing, and computational phonology. Section~\ref{sec:phonology} formalizes the ternary distinctive-feature representation, the three phone-distance metrics, and the generalized weighted edit distance. Section~\ref{sec:architecture} details the unified architecture, including keyboard layouts with Unicode NFD diacritic tolerance, the six language-specific G2P frontends, and the two-stage candidate-generation and scoring algorithm. Section~\ref{sec:setup} describes the corpora, benchmarks, clean $50/50$ non-word split construction, and index footprints across all six languages. Section~\ref{sec:results} presents ablation studies, full-benchmark comparisons against Aspell, Hunspell, and Norvig, cross-lingual transfer and per-language tuning, G2P dictionary ablations, qualitative error analysis, and limitations. Finally, Section~\ref{sec:conclusion} summarizes our findings and outlines future directions.

\section{Related Work}
\label{sec:related}

\subsection{Orthographic Edit Distance and Noisy-Channel Models}
Early work by \citet{damerau1964technique} established that more than $80\%$ of typographical errors in typed text correspond to a single character insertion, deletion, substitution, or adjacent transposition. Building on this observation, \citet{church1991probability} formulated spelling correction as a Bayesian noisy-channel problem, ranking each candidate word $w$ for an observed misspelling $x$ by $P(w)\,P(x \mid w)$, where $P(w)$ is a corpus unigram prior and $P(x \mid w)$ models character-level confusion probabilities estimated from large corpora. \citet{brill2000improved} extended the noisy-channel error model from single-character edits to arbitrary substring substitutions ($a \rightarrow b$), allowing multi-character orthographic patterns such as \texttt{ph} $\rightarrow$ \texttt{f} or \texttt{ent} $\rightarrow$ \texttt{ant} to be learned from paired misspelling data. \citet{norvig2007spell} popularized a clean, parameter-free baseline that generates all strings within Damerau--Levenshtein distance $1$ or $2$ of $x$ and selects the candidate maximizing unigram corpus probability.

To accelerate candidate retrieval on large vocabularies, \citet{garbe2012symspell} introduced the Symmetric Delete (\textit{SymSpell}) algorithm: instead of generating deletions, insertions, substitutions, and transpositions of the query at lookup time, SymSpell pre-indexes all $0$--$k$-character deletions of every dictionary word and intersects them only with $0$--$k$-character deletions of the query. LipSpell adopts SymSpell-style $0$--$2$-character orthographic deletions as its first retrieval stage, but pairs it with a second phonetic-code stage so that multi-edit phonetic misspellings ($d_{\text{orth}} > 2$) still enter the candidate pool.

\subsection{Phonetic Hashing and Pronunciation-Aware Systems}
Classical phonetic indexing algorithms such as Soundex \citep{russell1918soundex}, Metaphone \citep{philips1990metaphone}, and Double Metaphone \citep{philips2000double} (which adds secondary encodings for non-English borrowings) map letter sequences into coarse, language-specific phonetic codes designed primarily for English proper names and Latin/ASCII orthography. Widely deployed open-source spell checkers illustrate the limitations of hard-coded language heuristics: GNU Aspell \citep{atkinson_aspell} combines edit distance with Metaphone-style keys and hand-authored per-language phonetic replacement tables, while Hunspell \citep{nemeth_hunspell} emphasizes morphological affixation rules and orthographic edit suggestions. Because these heuristics are coupled to specific alphabets or hand-crafted phonetic rules, their accuracy varies sharply across scripts and morphologically rich languages (e.g., GNU Aspell's Turkish dictionary covers only static uninflected forms, whereas Hunspell lacks corpus frequency priors to break ties among unvocalized Hebrew roots).

To model pronunciation errors in a probabilistic framework, \citet{toutanova2002pronunciation} composed a letter-to-phone and phone-to-letter noisy channel with an orthographic channel, demonstrating substantial gains over purely orthographic models on English. However, training character-to-phone confusion tables requires aligned misspelling corpora for each target language. LipSpell avoids per-language confusion-matrix training by deriving phone substitution costs directly from universal articulatory distinctive features.

\subsection{Distinctive Features and Cross-Lingual Computational Phonology}
In generative phonology \citep{chomsky1968sound}, speech sounds are decomposed into binary or ternary articulatory and acoustic distinctive features (such as $[\pm\text{syllabic}]$, $[\pm\text{consonantal}]$, $[\pm\text{voice}]$, and place/manner specifications). \citet{hulden2012distinctive} compiled a compact $24$-feature ternary IPA table ($\{-1, 0, +1\}$) within the \texttt{foma} finite-state toolkit, which we adopt and extend with Portuguese palatal consonants.

In computational phonology, Kondrak's ALINE algorithm \citep{kondrak2003phonetic} adapts sequence-alignment dynamic programming from bioinformatics (analogous to Needleman--Wunsch and Smith--Waterman) to align phonetic transcriptions and measure cross-linguistic word similarity for cognate identification. ALINE represents each phoneme as a vector of multivalued and binary articulatory features (place, manner, voicing, vowel height, and backness) weighted by feature salience, penalizing mismatches, skips (insertions/deletions), and $1$-to-$2$ expansions/compressions (e.g., aligning a diphthong with a monophthong) to output a normalized similarity score. LipSpell differs from ALINE in three key respects: (1) rather than standalone cognate similarity, LipSpell integrates phonetic alignment into an end-to-end spelling-correction engine alongside orthographic edit distance, physical keyboard geometry, corpus unigram priors, and two-stage candidate indexing; (2) whereas ALINE employs a distinct mixed inventory of multivalued and binary articulatory features (overlapping only partially with SPE features) and evaluates feature differences inside each DP step, LipSpell uses Hulden's uniform $24$-feature ternary SPE-style representation $\{-1,0,+1\}^{24}$ and precomputes an $O(1)$-lookup $|P|\times|P|$ substitution matrix $M$, enabling direct comparison of $L_1$, salience-$L_1$, and cosine metrics; and (3) LipSpell uses $1$-to-$1$ generalized weighted Damerau--Levenshtein DP with adjacent phone transpositions ($\tau=0.45$) rather than $1$-to-$2$ expansions/compressions.

Later resources such as PanPhon \citep{mortensen2016panphon} and Epitran \citep{mortensen2018epitran} provide broad-coverage IPA feature vectors and multilingual rule-based G2P transducers, respectively. Expanding our earlier English prototype \citep{araujo2018phonetic}, LipSpell precomputes feature-based phone substitution matrices in NumPy, unifies them with physical keyboard geometry and corpus frequency priors, and evaluates transfer across six languages and three writing systems.

\subsection{Multilingual and L2 Spelling Benchmarks}
Standard English evaluations rely on historical collections such as Birkbeck \citep{mitton1985birkbeck} (including the Holbrook schoolchild subset), Norvig's two test sets, and Aspell's benchmark list \citep{norvig2007spell}. More recently, \citet{flor2019toefl} introduced TOEFL-Spell, a large corpus of non-native L2 English learner misspellings where phonetic approximations and vowel confusions are prevalent. For Brazilian Portuguese, \citet{bertaglia2016enelvo} released normalized user-generated corpora and clean lexicons (\texttt{Enelvo}), while pronunciation resources include Corpus ABG \citep{benevides2017abg}, \texttt{ipa-dict} \citep{opendictipadict}, and WikiPron \citep{lee2020wikipron}. Cross-lingual benchmarks in Russian (\texttt{RUSpellRU}), Hebrew, Turkish, and French paired with OpenSubtitles frequency lists \citep{lison2016opensubtitles} allow systematic testing of orthographic depth---from shallow Turkish to deep/opaque French and unvocalized Hebrew Abjad.

\subsection{Neural and Context-Aware Spelling Correction}
A parallel line of research addresses spelling and grammatical error correction using sequence-to-sequence neural networks, masked language models, and large language models, which excel at context-dependent real-word confusions (such as \textit{their}/\textit{there} or \textit{mas}/\textit{mais}). Because LipSpell targets lightweight, interpretable isolated-word non-word correction with zero neural training corpora, it trades sentence-level contextual disambiguation for fast, deterministic CPU inference and transparent phonetic-feature scoring, making it complementary to heavier neural rerankers.

\section{Phonological Principles and Feature-Based Metrics}
\label{sec:phonology}
In generative phonology---most influentially formalized in Chomsky and Halle's \emph{The Sound Pattern of English} (SPE) \citep{chomsky1968sound}---phonemes are not treated as atomic, indivisible units. Instead, each speech sound is decomposed into a simultaneous bundle (feature matrix) of subsegmental \emph{distinctive features}, the smallest building blocks of phonological structure used to contrast one phoneme from another. Shifting away from earlier primarily acoustic feature inventories (such as Jakobson's), SPE grounded distinctive features in articulatory definitions and binary specifications ($+$ for feature presence, $-$ for absence) operating over abstract underlying representations mapped to surface phonetic forms by ordered rules. Crucially, distinctive features define \emph{natural classes}---groups of sounds sharing specific properties (such as $[+\text{voice}]$ or $[+\text{consonantal}]$) that pattern together in phonological processes---and provide a measure of formal rule simplicity, since generalizations requiring fewer feature specifications are simpler and broader.

To embed the full IPA segment inventory $P$ in a uniform vector space for fast pairwise comparison, we represent each segment $p \in P$ as a $24$-dimensional ternary vector $\mathbf{v}_p \in \{-1, 0, +1\}^{24}$ from the Hulden IPA feature table \citep{hulden2012distinctive}, extended with the palatal nasal /\textipa{J}/ (\texttt{nh}) and palatal lateral /\textturny/ (\texttt{lh}) required for Portuguese. Values $+1$ and $-1$ correspond directly to SPE's binary $+$ and $-$ specifications, while $0$ denotes an unspecified (non-applicable) feature (e.g., vowel height features \texttt{hi}, \texttt{lo}, \texttt{fr}, \texttt{bk}, and \texttt{tns} take value $0$ for consonants, whereas consonant place features such as \texttt{labdent}, \texttt{ant}, and \texttt{dist} take value $0$ for vowels). Consequently, phonetically similar sounds (e.g., /p/ and /b/, differing only in $[-\text{voi}]$ vs.\ $[+\text{voi}]$) sit at minimal distance in $\{-1,0,+1\}^{24}$, whereas sounds from distinct natural classes differ across multiple specifications.

We precompute the $|P|\times|P|$ substitution matrix $M$ and indel vector $\mathbf{d}_{\text{indel}} \in \mathbb{R}^{|P|}$ once at initialization, making every dynamic programming (DP) transition an expected $O(1)$ array lookup. For any two IPA segments $u, v \in P$ with corresponding $24$-dimensional feature vectors $\mathbf{v}_u, \mathbf{v}_v \in \{-1, 0, +1\}^{24}$ (whose $k$-th coordinates are $v_{u,k}$ and $v_{v,k}$), we compare three distance metrics for the matrix entry $M_{u,v}$:
\begin{enumerate}
    \item \textbf{Uniform $\ell_1$\footnote{Throughout, $\ell_1$ denotes the normalized Manhattan ($\ell_1$-norm) distance between ternary phonological feature vectors---distinct from first language (L1)---and \emph{salience} denotes fixed phonological feature weights $w_k$ inspired by Kondrak's ALINE \citep{kondrak2003phonetic}, rather than machine-learning feature-attribution saliency.}:}
    \begin{equation}
        M_{u,v}^{\ell_1} = \frac{1}{2F}\sum_{k=1}^{F} |v_{u,k} - v_{v,k}|, \quad F=24.
    \end{equation}
    Because unspecified ($0$) features contribute $0$ to the numerator while dividing by the fixed denominator $2F=48$, segments sharing many $0$s appear spuriously close under $\ell_1$. For instance, two consonants sharing $5$ unspecified vowel features have their effective active feature count reduced, compressing consonant--consonant distances relative to vowel--vowel pairs.
    \item \textbf{Salience-weighted $\ell_1$:} Inspired by feature weighting in ALINE \citep{kondrak2003phonetic},
    \begin{equation}
        M_{u,v}^{\text{sal}} = \frac{\sum_{k=1}^{F} w_k |v_{u,k} - v_{v,k}|}{2\sum_{k=1}^{F} w_k},
    \end{equation}
    where fixed phonological weights $\mathbf{w}$ penalize major-class and primary-place mismatches more heavily than voicing or tense/lax vowel shifts that frequently occur in cognitive misspellings. Specifically, our $24$-feature salience vector assigns: $w_{\text{syll}}=2.0$; $w_{\text{cons}}=1.8$; $1.5$ to \texttt{son}, \texttt{cont}, \texttt{nas}, \texttt{lab}, \texttt{cor}, and \texttt{dor}; $1.2$ to \texttt{approx}; $1.0$ to \texttt{delrel}, \texttt{labdent}, \texttt{strid}, and \texttt{lat}; $0.8$ to \texttt{round}, \texttt{ant}, \texttt{hi}, \texttt{lo}, \texttt{fr}, and \texttt{bk}; $0.6$ to \texttt{voi} and \texttt{dist}; and $0.5$ to \texttt{spread}, \texttt{constr}, and \texttt{tns}. Thus, a voicing slip (/v/ $\leftrightarrow$ /f/) or tense/lax confusion (/\textsci/ $\leftrightarrow$ /i/) incurs a small substitution penalty, whereas replacing a vowel with a stop consonant incurs a large penalty.
    \item \textbf{Cosine distance:}
    \begin{equation}
        M_{u,v}^{\cos} = \frac{1}{2}\left(1 - \frac{\mathbf{v}_u \cdot \mathbf{v}_v}{\lVert \mathbf{v}_u \rVert_2 \, \lVert \mathbf{v}_v \rVert_2}\right) \in [0, 1].
    \end{equation}
    Because $\mathbf{v}_p \in \{-1, 0, +1\}^{24}$, cosine similarity lies in $[-1, 1]$, so $1-\cos \in [0, 2]$ and the factor $\frac{1}{2}$ rescales to $[0, 1]$. In our $45$-segment table, consonants average $18.25 \pm 1.50$ active (non-zero) features ($n=28$) versus $19.94 \pm 0.87$ for vowels ($n=17$). Normalizing by the $\ell_2$ norm $\lVert \mathbf{v}_p \rVert_2$ equalizes vector magnitudes across segments with different numbers of unspecified ($0$) entries, preventing sparse consonants from appearing artificially closer than vowels simply due to shared zeros, though non-applicable $0$s can still yield imperfect cross-class similarities when a consonant and vowel happen to agree on secondary manner features.
\end{enumerate}

For insertion and deletion costs, under $\ell_1$ ($w_k=1$) and salience-weighted $\ell_1$, $\mathbf{d}_{\text{indel}}(p)$ is proportional to active feature density and clipped to $[0.35, 0.75]$:
\begin{equation}
    \mathbf{d}_{\text{indel}}(p) = \operatorname{clip}\!\left(1.4 \cdot \frac{\sum_{k=1}^{F} w_k |v_{p,k}|}{2\sum_{k=1}^{F} w_k},\; 0.35,\; 0.75\right),
\end{equation}
using identical insertion and deletion costs (so segments with fewer active features, such as /h/ or /\textschwa/, are cheaper to insert or delete than heavily specified obstruents). Under cosine distance, $\mathbf{d}_{\text{indel}}(p)=0.5$ uniformly. Table~\ref{tab:salience} lists the complete $24$-feature salience weights grouped by phonological dimension. Illustrative substitution costs confirm the geometric behavior of these computational feature vectors: voicing (/p/--/b/: $\ell_1=0.0417$, salience $=0.0235$, cosine $=0.0625$) and tense/lax vowel shifts (/i/--/\textsci/: $\ell_1=0.0417$, salience $=0.0196$, cosine $=0.0500$) are much cheaper than place changes (/p/--/t/: $\ell_1=0.1458$, salience $=0.1647$, cosine $=0.1559$) or vowel height/backness shifts (/i/--/u/: $\ell_1=0.1667$, salience $=0.1529$, cosine $=0.2000$), whereas cross-class consonant--vowel substitution (/p/--/i/) drops below $0.5$ under uniform $\ell_1$ ($0.4583$) due to shared unspecified zeros while remaining penalized above $0.5$ under salience ($0.5471$) and cosine ($0.5280$). Because $M$ and $\mathbf{d}_{\text{indel}}$ assign non-uniform costs (and cosine distance does not strictly satisfy the triangle inequality), the resulting phone alignment defines a \emph{weighted Optimal String Alignment (OSA) distance} rather than a strict metric, which remains exact under dynamic programming.

\begin{table}[ht]
\centering
\small
\caption{Complete $24$-feature salience weights $\mathbf{w}$ used in salience-weighted $L_1$ phone distance.}
\label{tab:salience}
\begin{tabular}{lll}
\toprule
Phonological Group & Features & Weight $w_k$ \\
\midrule
Major class   & \texttt{syll} & $2.0$ \\
              & \texttt{cons} & $1.8$ \\
              & \texttt{son} & $1.5$ \\
Manner        & \texttt{cont}, \texttt{nas} & $1.5$ \\
              & \texttt{approx} & $1.2$ \\
              & \texttt{delrel}, \texttt{strid}, \texttt{lat} & $1.0$ \\
Primary place & \texttt{lab}, \texttt{cor}, \texttt{dor} & $1.5$ \\
              & \texttt{labdent} & $1.0$ \\
              & \texttt{ant}, \texttt{round} & $0.8$ \\
              & \texttt{dist} & $0.6$ \\
Vowel space   & \texttt{hi}, \texttt{lo}, \texttt{fr}, \texttt{bk} & $0.8$ \\
              & \texttt{tns} & $0.5$ \\
Laryngeal     & \texttt{voi} & $0.6$ \\
              & \texttt{spread}, \texttt{constr} & $0.5$ \\
\bottomrule
\end{tabular}
\end{table}

\section{Unified Architecture}
\label{sec:architecture}

\subsection{Keyboard Geometry, Diacritic Tolerance, and Weak-Character Indels}
Physical key coordinates are modeled in two-dimensional row-staggered key-width units with horizontal row offsets $(0.0, 0.5, 0.75, 1.25)$ across six standard layouts: QWERTY (English), ABNT2 (Brazilian Portuguese, adding \texttt{\c{c}} and dead-key accents), JCUKEN (Russian Cyrillic), SI-1452 (Israeli Hebrew), Turkish-Q (adding \texttt{\u{g}}, \texttt{\"{u}}, \texttt{\c{s}}, \texttt{\i}, \texttt{\"{o}}, \texttt{\c{c}}), and AZERTY (French). For every layout, we precompute a $|K|\times|K|$ symmetric substitution matrix by clamping Euclidean key distance at a saturation radius of $3.0$ key widths and normalizing to $[0, 1]$, so an adjacent-key slip (distance $\approx 1.0$) incurs a substitution cost near $0.33$.

To handle omitted or swapped diacritics in languages with accented Latin orthographies without penalizing them as distant keystrokes, \texttt{KeyboardLayout.distance} applies Unicode canonical decomposition (NFD) to both characters, separating base letters from combining diacritical marks (e.g., \textit{\'a}$\rightarrow$\textit{a}+\,\textasciiacute). Whenever two characters share the same base letter and at least one carries a combining mark (such as \textit{a}/\textit{\'a}, \textit{c}/\textit{\c{c}}, \textit{e}/\textit{\^e}, or \textit{s}/\textit{\c{s}}), the keyboard substitution cost is fixed to $0.15$ regardless of dead-key layout position. Inside the keyboard-weighted Damerau--Levenshtein DP, accidental insertion or deletion adjacent to a neighboring character $c_{\text{adj}}$ receives a reduced indel cost $\min(0.7, 0.4 + 0.6\,d_{\text{kb}}(c, c_{\text{adj}}))$, capturing finger-slip double strikes and adjacent-key insertions. Finally, for consonantal abjad scripts such as Hebrew where \textit{matres lectionis} (\textit{alef}/\textit{vav}/\textit{yod}) are optionally inserted or omitted as vowel markers without altering consonant identity, LipSpell supports per-character indel overrides ($0.25$ instead of $0.7$).

\subsection{Language-Specific G2P Frontends and IPA Tokenization}
For the grapheme-to-phoneme (G2P) mapping $\phi$, which converts a written word $w$ into its IPA phone sequence $\phi(w)$, only English optionally queries a pronunciation dictionary (\texttt{CMUdict}); all five other languages operate with compact, deterministic, purely rule-based G2P frontends requiring zero external pronunciation lexicons:
\begin{itemize}
    \item \textbf{English (\texttt{EnglishG2P}):} Looks up known words in CMUdict (\texttt{cmudict-0.7b}) \citep{weide1998cmu} after stripping lexical stress digits (\texttt{0/1/2}) and mapping ARPAbet phones to IPA, and falls back to an ordered rule-based G2P for misspelled or out-of-vocabulary (OOV) inputs. The rule set handles common English digraphs (\texttt{ph}$\rightarrow$/f/, \texttt{sh}$\rightarrow$/\textesh/, \texttt{ch}$\rightarrow$/\textteshlig/, \texttt{th}$\rightarrow$/\texttheta/, \texttt{tion}$\rightarrow$/\textesh\textschwa n/) and CVCe silent-\texttt{e} vowel lengthening (\textit{nite}$\rightarrow$/na\textsci t/, matching \textit{night}).
    \item \textbf{Brazilian Portuguese (\texttt{PortugueseG2P}):} A deterministic phonemic approximation of Southeastern Brazilian Portuguese comprising ordered multi-character, contextual, and single-character rules: palatal digraphs (\texttt{ch}$\rightarrow$/\textesh/, \texttt{lh}$\rightarrow$/\textturny/, \texttt{nh}$\rightarrow$/\textipa{J}/), velar digraphs (\texttt{qu}$\rightarrow$/k/, \texttt{gu}$\rightarrow$/\textipa{g}/ applied unconditionally), contextual sibilants and rhotics (\texttt{sc}/\texttt{xc} before \texttt{e}/\texttt{i}$\rightarrow$/s/, \texttt{c}/\texttt{g} before \texttt{e}/\texttt{i}$\rightarrow$/s, \textyogh/, intervocalic \texttt{s}$\rightarrow$/z/, word-initial \texttt{r} and \texttt{rr}$\rightarrow$/h/), silent \texttt{h}, and accented vowels. Nasal vowels and diphthongs are deliberately folded onto their oral/central counterparts (\texttt{\^a/\~a}$\rightarrow$/\textturna/, \texttt{\^o/\~o}$\rightarrow$/o/, \texttt{\~ao}$\rightarrow$/a\textupsilon/, \texttt{\~oe}$\rightarrow$/o\textsci/, \texttt{\~ae}$\rightarrow$/a\textsci/) so that omitted tildes or circumflexes do not incur artificial phonetic penalties.
    \item \textbf{Russian (\texttt{RussianG2P}):} A rule-based Cyrillic-to-IPA transducer mapping iotated vowels (\textit{e/yo/yu/ya} $\rightarrow$ /je, jo, ju, ja/), word-final genitive suffixes (\textit{-ogo/-ego} $\rightarrow$ /ovo/, /jevo/), and word-final obstruent devoicing (\textit{b, v, g, d, zh, z} $\rightarrow$ /p, f, k, t, \textesh, s/), while stripping hard/soft signs (\textit{tvjordyj/mjagkij znak} $\rightarrow \emptyset$; consonant palatalization is not modeled separately).
    \item \textbf{Hebrew (\texttt{HebrewG2P}):} An unvocalized abjad letter-to-IPA mapping over the $22$ Hebrew consonants and $5$ word-final forms (\textit{sofit}), neutralizing historical pharyngeal/emphatic mergers common in Modern Hebrew (\textit{alef}/\textit{ayin}$\rightarrow$/\textglotstop/, \textit{tet}/\textit{tav}$\rightarrow$/t/, \textit{kaf}/\textit{qof}$\rightarrow$/k/, while mapping \textit{bet}$\rightarrow$/b/ and \textit{vav}$\rightarrow$/v/).
    \item \textbf{Turkish (\texttt{TurkishG2P}):} A shallow $1$-to-$1$ grapheme-to-IPA mapping paired with Turkish-aware case folding (\texttt{\.{I}}$\rightarrow$\texttt{i}, \texttt{I}$\rightarrow$\texttt{\i}) so dotted and dotless \texttt{i}/\texttt{\i} are never conflated prior to indexing, mapping \texttt{c}$\rightarrow$/\textdyoghlig/, \texttt{\c{c}}$\rightarrow$/\textteshlig/, \texttt{\c{s}}$\rightarrow$/\textesh/, \texttt{\i}$\rightarrow$/\textturnm/, and eliding the silent lengthening \textit{yumuşak ge} (\texttt{\u{g}}).
    \item \textbf{French (\texttt{FrenchG2P}):} A rule-based approximation of French orthography-to-IPA including silent word-final consonant (\texttt{s, t, d, p, x, z}) and silent-\texttt{e} stripping, multi-graphs (\texttt{eau/au}$\rightarrow$/o/, \texttt{tion}$\rightarrow$/sj\textopeno n/, \texttt{ch}$\rightarrow$/\textesh/, \texttt{gn}$\rightarrow$/\textipa{J}/, \texttt{ou}$\rightarrow$/u/), nasal vowel sequences (\texttt{an/am/en/em}$\rightarrow$/\textscripta/, \texttt{on/om}$\rightarrow$/o/, \texttt{in/im/un}$\rightarrow$/\textepsilon/) before consonants or word boundaries, and doubled-consonant collapsing.
\end{itemize}
Table~\ref{tab:pt_g2p} summarizes the ordered deterministic rules of \texttt{PortugueseG2P} as a representative non-English rule-based frontend, and \ref{app:g2p} provides the complete rule specifications for all six languages. At tokenization time, IPA strings emitted by any frontend are segmented greedily longest-first against $P$: two-character affricates /\textteshlig/ and /\textdyoghlig/ are tokenized as single segments in $P$, diphthongs (e.g., /a\textsci/, /a\textupsilon/) are tokenized as two adjacent vowel segments, and any unknown symbols or suprasegmental diacritics are skipped.

\begin{table}[ht]
\centering
\scriptsize
\setlength{\tabcolsep}{3.5pt}
\caption{Ordered deterministic rules in \texttt{PortugueseG2P} (priority: contextual/multi-graph $\rightarrow$ single grapheme; $1$-to-$1$ consonants omitted).}
\label{tab:pt_g2p}
\begin{tabular}{lp{5.8cm}ll}
\toprule
Category & Pattern / Rule & IPA Output & Examples \\
\midrule
Consonant digraphs & \texttt{ch}$\rightarrow$/\textesh/, \texttt{lh}$\rightarrow$/\textturny/, \texttt{nh}$\rightarrow$/\textipa{J}/, \texttt{rr}$\rightarrow$/h/, \texttt{ss}$\rightarrow$/s/ & /\textesh, \textturny, \textipa{J}, h, s/ & \textit{chave, milho, banho, carro} \\
Velar digraphs & \texttt{qu}$\rightarrow$/k/, \texttt{gu}$\rightarrow$/\textipa{g}/ (unconditional; omits /kw, gw/) & /k, \textipa{g}/ & \textit{queijo, guerra} \\
Diphthongs (oralized) & \texttt{\~ao}$\rightarrow$/a\textupsilon/, \texttt{\~oe}$\rightarrow$/o\textsci/, \texttt{\~ae}$\rightarrow$/a\textsci/, \texttt{ou}$\rightarrow$/o/, \texttt{ei}$\rightarrow$/e\textsci/, \texttt{ai}$\rightarrow$/a\textsci/, \texttt{oi}$\rightarrow$/\textopeno\textsci/, \texttt{au}$\rightarrow$/a\textupsilon/ & /a\textupsilon, o\textsci, a\textsci, o, e\textsci, a\textsci, \textopeno\textsci, a\textupsilon/ & \textit{cora\c{c}\~ao, p\~oe, m\~ae, ouro} \\
Contextual rules & \texttt{sc/xc}+\texttt{e,i}$\rightarrow$/s/; \texttt{c}+\texttt{e,i}$\rightarrow$/s/; \texttt{g}+\texttt{e,i}$\rightarrow$/\textyogh/; \texttt{V\_s\_V}$\rightarrow$/z/; \texttt{\#r}$\rightarrow$/h/ & /s, \textyogh, z, h/ & \textit{nascer, cedo, gelo, casa, rato} \\
Vowels \& consonants & \texttt{\'a/\`a}$\rightarrow$/a/, \texttt{\^a/\~a}$\rightarrow$/\textturna/, \texttt{\'e}$\rightarrow$/\textepsilon/, \texttt{\^e}$\rightarrow$/e/, \texttt{\'o}$\rightarrow$/\textopeno/, \texttt{\^o/\~o}$\rightarrow$/o/, \texttt{\c{c}}$\rightarrow$/s/, \texttt{j}$\rightarrow$/\textyogh/, \texttt{x}$\rightarrow$/\textesh/, \texttt{h}$\rightarrow\emptyset$ & /a, \textturna, \textepsilon, e, \textopeno, o, s, \textyogh, \textesh/ & \textit{caf\'e, av\^o, ca\c{c}a, hora} \\
\bottomrule
\end{tabular}
\end{table}

\subsection{Two-Stage Candidate Generation and Generalized Scoring}
\label{sec:indexing}
To avoid scanning $75\text{k}$--$777\text{k}$ vocabulary entries on each query, LipSpell builds a two-stage inverted index at startup:
\begin{enumerate}
    \item \textbf{Orthographic symmetric-delete index:} Following \citet{garbe2012symspell}, for each vocabulary word $w \in V$ we pre-index all strings obtained by deleting $0$ to $2$ characters from $w$. At query time, intersecting $0$--$2$-character deletions of $x$ with the index recovers every candidate within Damerau--Levenshtein distance $d_{\text{orth}}(x, w) \le 2$ in $O(1)$ hash lookups per deletion.
    \item \textbf{Seven-class phonetic-code index:} Each vocabulary word's IPA phone sequence $\phi(w)$ is mapped onto a coarse seven-class phonological code via the mutually exclusive priority order $\texttt{V} > \texttt{N} > \texttt{L} > \texttt{S} > \texttt{P} > \texttt{T} > \texttt{K}$ (Table~\ref{tab:skel_classes}), collapsing consecutive identical classes and indexing both the exact code and its $1$-symbol deletions to tolerate an omitted or inserted phonological class.\footnote{In the codebase, this seven-class phonetic key is implemented as \texttt{skeleton}, denoting a Soundex-like broad phonological equivalence partition \citep{kukich1992techniques} rather than a vocal-tract articulatory model.} Posting lists per key are sorted by descending corpus frequency and capped at $300$ hits per query to bound DP alignment cost on populated keys. For example, \textit{cemetery} (/\text{s\textepsilon m\textschwa t\textepsilon\textturnr i}/) and \textit{semetary} (/\text{s\textepsilon m\textschwa t\textepsilon\textturnr i}/) both map to \texttt{SVNVTVLV} (identical code), while an omitted vowel or liquid in a distorted spelling (e.g., \texttt{SVNTVLV}) still collides via the $1$-symbol deletion index.
\end{enumerate}

\begin{table}[ht]
\centering
\small
\caption{Seven-class phonetic code partition $g: P \rightarrow \{\texttt{V},\texttt{N},\texttt{L},\texttt{S},\texttt{P},\texttt{T},\texttt{K}\}$ evaluated in priority order via $v_{p,f} > 0$ (treating unspecified $0$ and $-1$ alike as false).}
\label{tab:skel_classes}
\begin{tabular}{clll}
\toprule
Code & Phonological Class & Feature Criterion ($v_{p,f} > 0$) & Representative IPA Segments \\
\midrule
\texttt{V} & Syllabic / Vowel     & \texttt{syll} & /a, e, \textepsilon, i, \textsci, o, \textopeno, u, \textschwa, \textturna/ \\
\texttt{N} & Nasal                & \texttt{nas} & /m, n, \textipa{N}, \textipa{J}/ \\
\texttt{L} & Liquid / Approximant & \texttt{approx} or \texttt{lat} & /l, \textturny, \textfishhookr, \textturnr, j, w/ \\
\texttt{S} & Sibilant / Affricate & \texttt{strid} & /s, z, \textesh, \textyogh, \textteshlig, \textdyoghlig/ \\
\texttt{P} & Labial               & \texttt{lab} or \texttt{labdent} & /p, b, f, v/ \\
\texttt{T} & Coronal              & \texttt{cor} & /t, d, \texttheta, \dh/ \\
\texttt{K} & Dorsal / Glottal     & Otherwise (fallback) & /k, \textipa{g}, x, h, \textglotstop/ \\
\bottomrule
\end{tabular}
\end{table}

For query $x$ and each retrieved candidate $w$, we compute weighted Optimal String Alignment (OSA) distances $d_{\text{orth}}, d_{\text{kb}}, d_{\text{phon}}$ (restricted Damerau--Levenshtein allowing at most one edit per substring, using adjacent-transposition cost $\tau=0.45$ in keyboard and phonetic DP) and combine them with a Laplace-smoothed unigram log-prior $\log P(w) = \log(c(w)+0.5) - \log\!\left(\sum_{v\in V} c(v) + 0.5|V|\right)$ over the merged lowercase vocabulary $V$ (preserving diacritics):
\begin{equation}
    S(w \mid x) = \lambda_{\text{freq}} \log P(w) - \lambda_{\text{orth}} d_{\text{orth}}(x,w) - \lambda_{\text{kb}} d_{\text{kb}}(x,w) - \lambda_{\text{phon}} d_{\text{phon}}(\phi(x),\phi(w)).
\end{equation}
This is a \emph{noisy-channel-inspired scoring function}: the three edit distances are correlated rather than independent log-likelihoods, and $(\lambda_{\text{freq}}, \lambda_{\text{orth}}, \lambda_{\text{kb}}, \lambda_{\text{phon}}) = (1.0, 4.0, 1.5, 4.0)$ were tuned via grid search and coordinate refinement on an independent $800$-pair Birkbeck dev sample ($77.47\%$ dev Top-1 on the $586$ pairs in the candidate pool). Ties are broken deterministically by candidate string order. Algorithm~\ref{alg:lipspell} summarizes query-time retrieval and ranking, and Algorithm~\ref{alg:weighted_dp} specifies the weighted OSA dynamic program used to compute $d_{\text{kb}}$ (with context-dependent adjacent-key indels from Section~4.1) and $d_{\text{phon}}$ (with per-phone $\mathbf{d}_{\text{indel}}$ from Section~3). We refer to this four-signal model (using the cosine phone metric unless noted) as \textbf{LipSpell} to distinguish it from single-channel baselines (\texttt{Norvig}, \texttt{Keyboard}, \texttt{Phonetic-only}), while the scoring framework itself is language-independent.

\begin{algorithm}[ht]
\caption{Two-Stage Candidate Retrieval and Four-Signal Scoring in LipSpell}
\label{alg:lipspell}
\begin{algorithmic}[1]
\Require Query $x$, vocabulary $V$, deletion index $\mathcal{I}_{\text{del}}$, phonetic-code index $\mathcal{I}_{\text{code}}$, weights $\boldsymbol{\lambda}$, cap $H_{\max}=300$
\Ensure Top-$k$ ranked candidate corrections
\State $w_{\text{low}} \gets \text{lowercase}(x)$
\If{$w_{\text{low}} \in V$} \Return $[w_{\text{low}}]$ \Comment{Known in-vocabulary word}
\EndIf
\State $\mathcal{C}_{\text{orth}} \gets \bigcup_{s \in \text{Deletes}(w_{\text{low}}, 2)} \mathcal{I}_{\text{del}}[s]$ \Comment{Stage 1: SymDelete ($\le 2$ edits)}
\State $\mathbf{p}_x \gets \text{Tokenize}(\phi(w_{\text{low}}))$; \quad $\kappa_x \gets \text{PhoneticCode}(\mathbf{p}_x)$
\State $\mathcal{C}_{\text{phon}} \gets \emptyset$
\For{$s \in \text{Deletes}(\kappa_x, 1)$} \Comment{Stage 2: 7-class code + 1-delete}
    \For{$c \in \mathcal{I}_{\text{code}}[s]$} \Comment{Pre-sorted by descending corpus frequency}
        \State $\mathcal{C}_{\text{phon}} \gets \mathcal{C}_{\text{phon}} \cup \{c\}$
        \If{$|\mathcal{C}_{\text{phon}}| \ge H_{\max}$} \textbf{break} \EndIf
    \EndFor
\EndFor
\State $\mathcal{C} \gets \mathcal{C}_{\text{orth}} \cup \mathcal{C}_{\text{phon}}$
\For{$w \in \mathcal{C}$}
    \State $S(w \mid x) \gets \lambda_{\text{freq}} \log P(w) - \lambda_{\text{orth}} d_{\text{orth}}(w_{\text{low}}, w) - \lambda_{\text{kb}} d_{\text{kb}}(w_{\text{low}}, w) - \lambda_{\text{phon}} d_{\text{phon}}(\mathbf{p}_x, \mathbf{p}_w)$
\EndFor
\State \Return Top-$k$ candidates in $\mathcal{C}$ ordered by $(-S(w \mid x), w)$
\end{algorithmic}
\end{algorithm}

\begin{algorithm}[ht]
\caption{Weighted Optimal String Alignment (OSA) Dynamic Program ($d_{\text{phon}}$ / $d_{\text{kb}}$)}
\label{alg:weighted_dp}
\begin{algorithmic}[1]
\Require Sequences $A = (a_1,\dots,a_m)$ and $B = (b_1,\dots,b_n)$, substitution cost $\text{sub}(a,b)$, position-aware indel costs $\text{del}(a_i, i, A)$ and $\text{ins}(b_j, j, B)$ (per-phone $\mathbf{d}_{\text{indel}}$ for $d_{\text{phon}}$; context-dependent adjacent-key costs from Sec.~4.1 for $d_{\text{kb}}$), transposition cost $\tau = 0.45$
\Ensure Weighted OSA distance $D[m, n]$
\State $D[0, 0] \gets 0$
\For{$i = 1 \dots m$} $D[i, 0] \gets D[i-1, 0] + \text{del}(a_i, i, A)$ \EndFor
\For{$j = 1 \dots n$} $D[0, j] \gets D[0, j-1] + \text{ins}(b_j, j, B)$ \EndFor
\For{$i = 1 \dots m$}
    \For{$j = 1 \dots n$}
        \State $c_{\text{sub}} \gets 0$ \textbf{if} $a_i = b_j$ \textbf{else} $\text{sub}(a_i, b_j)$
        \State $D[i, j] \gets \min\big(D[i-1, j] + \text{del}(a_i, i, A),\; D[i, j-1] + \text{ins}(b_j, j, B),\; D[i-1, j-1] + c_{\text{sub}}\big)$
        \If{$i > 1$ \textbf{and} $j > 1$ \textbf{and} $a_i = b_{j-1}$ \textbf{and} $a_{i-1} = b_j$}
            \State $D[i, j] \gets \min\big(D[i, j],\; D[i-2, j-2] + \tau\big)$ \Comment{Adjacent transposition (OSA)}
        \EndIf
    \EndFor
\EndFor
\State \Return $D[m, n]$
\end{algorithmic}
\end{algorithm}

\section{Datasets and Experimental Setup}
\label{sec:setup}

\subsection{Lexicons and Evaluation Benchmarks}
All evaluation sets consist of deduplicated lowercase alphabetic misspelling--target pairs $(x, w^*)$ with $x \neq w^*$. For English and Brazilian Portuguese, we retain standard benchmark lexicons and frequency corpora from prior work \citep{norvig2007spell,bertaglia2016enelvo}, whereas for the four cross-lingual extensions (Russian, Hebrew, Turkish, and French) we pair Hunspell dictionary headwords (\texttt{index.dic} from \texttt{wooorm/dictionaries}, floor count $1$) with OpenSubtitles corpus unigram frequencies (\texttt{FrequencyWords}) \citep{lison2016opensubtitles}:
\begin{itemize}
    \item \textbf{English} ($|V|=78{,}393$): Word frequencies from Norvig's \texttt{big.txt} corpus \citep{norvig2007spell} merged with \texttt{/usr/share/dict/american-english} (floor $1$). We evaluate on six diverse benchmarks: \textbf{Norvig1} ($n=270$) and \textbf{Norvig2} ($n=400$), Norvig's two standard English test sets \citep{norvig2007spell}; \textbf{Aspell} ($n=516$), GNU Aspell's benchmark error list; \textbf{Wikipedia-EN} (first $300$ pairs), common English Wikipedia misspellings; \textbf{Holbrook} (first $300$ pairs in file order), British schoolchildren's misspellings from the Birkbeck corpus \citep{mitton1985birkbeck}; and \textbf{TOEFL-Spell} \citep{flor2019toefl} ($1{,}000$ unique L2 learner pairs from \texttt{Annotations.tsv} evaluated in full in Table~\ref{tab:english_cmp}, alongside its $n=489$ clean in-vocabulary non-word test split in Table~\ref{tab:multilingual}). We verified $0$ exact pair overlap between the $800$-pair Birkbeck dev set and all six evaluation sets (including Holbrook).
    \item \textbf{Brazilian Portuguese} ($|V|=240{,}040$): Word frequencies from Enelvo's \texttt{freq-cgu.txt} corpus merged with \texttt{unitex-full-clean} \citep{bertaglia2016enelvo} (floor $1$ for entries without explicit counts). We evaluate on the first $500$ alphabetical pairs, the complete A--Z dataset ($n=1{,}149$ single-word alphabetic pairs from \textit{Wikip\'edia: Lista de erros comuns}, eliminating alphabetical prefix bias), and the $50/50$ clean non-word split (test $n=499$).
    \item \textbf{Russian} ($|V|=358{,}624$): Hunspell \texttt{index.dic} merged with the top $250\text{k}$ OpenSubtitles frequencies; evaluated on the $50/50$ clean non-word split of \texttt{RUSpellRU} (test $n=321$).
    \item \textbf{Hebrew} ($|V|=339{,}791$): Top $250\text{k}$ OpenSubtitles frequencies restricted to Hunspell's \texttt{index.dic} to exclude noisy non-Hebrew tokens; evaluated on the $50/50$ clean non-word split of Wikipedia bot typos (test $n=122$).
    \item \textbf{Turkish} ($|V|=777{,}340$): Hunspell \texttt{index.dic} merged with the top $500\text{k}$ OpenSubtitles frequencies (reflecting agglutinative inflectional breadth); evaluated on the $50/50$ clean non-word split of Turkish Wikipedia typos ($1{,}000$-pair sample, test $n=420$).
    \item \textbf{French} ($|V|=74{,}544$): Top $250\text{k}$ OpenSubtitles frequencies restricted to Hunspell's \texttt{index.dic}; evaluated on the $50/50$ clean non-word split of French Wikipedia typos (test $n=106$).
\end{itemize}

\subsection{Clean 50/50 Split Construction and Parameter Tuning}
For the cross-lingual comparisons, \emph{50/50 clean non-word splits} are constructed deterministically from each language's raw pair list by retaining only single lowercase alphabetic pairs $(x, w^*)$ satisfying $x \notin V$, $w^* \in V$, and $x \neq w^*$, shuffling with \texttt{random.Random(42)}, and partitioning at index $\lfloor N/2 \rfloor$ into a development half ($n_{\text{dev}}=\lfloor N/2 \rfloor$) and a test half ($n_{\text{test}}=\lceil N/2 \rceil$, so odd-sized lists assign the extra pair to the test split, e.g., Russian $320/321$ and Hebrew $121/122$). The development half is used exclusively for unregularized grid search over the four scalar channel weights: $\lambda_{\text{freq}}=1$, $\lambda_{\text{orth}}\in\{2,4,6,8\}$, $\lambda_{\text{kb}}\in\{0,1,2,4\}$, and $\lambda_{\text{phon}}\in\{0,1,2,4,6\}$, selecting the tuple maximizing development Top-1 accuracy (and never to design or modify G2P rules, keyboard coordinates, or weak-character indel overrides, all of which were specified prior to test evaluation).

\subsection{Index Footprint and Statistical Evaluation Protocol}
Table~\ref{tab:index_stats} summarizes the startup index footprint across all six languages, reporting vocabulary size $|V|$, index construction time, number of hash keys in the symmetric-delete and seven-class phonetic-code tables, and isolated process peak resident RAM (RSS, measured via \texttt{ru\_maxrss}). Index memory scales primarily with the number of $0$--$2$-character deletion keys, which grows quadratically with word length: on English ($|V|=78\text{k}$), $2.37\text{M}$ delete keys require only $497.4$\,MiB RSS and $3.04$\,s build time, whereas on agglutinative Turkish ($|V|=777\text{k}$), $39.44\text{M}$ delete keys drive peak RSS to $8{,}650.6$\,MiB ($8.45$\,GiB / $8.65$\,GB) and $51.58$\,s build time. By contrast, the seven-class phonetic-code table remains compact across all six languages ($161\text{k}$--$2.14\text{M}$ keys). Because delete-key count scales quadratically with average word length, shorter-word unvocalized Hebrew ($|V|=339\text{k}$, $3.04\text{M}$ delete keys) builds faster ($9.12$\,s) than longer-word Portuguese ($|V|=240\text{k}$, $8.82\text{M}$ delete keys, $11.90$\,s).

All experiments ran on an Intel Core i7-13650HX ($14$\,GiB RAM, Ubuntu 26.04.1 LTS, Linux kernel 7.0, Python 3.14 + NumPy). We compare against Norvig \citep{norvig2007spell} (using the exact same vocabulary $V$ and frequency counts as LipSpell), Aspell 0.60.8.2 \citep{atkinson_aspell}, and Hunspell 1.7.2 \citep{nemeth_hunspell} (executed via UTF-8 pipe mode, evaluating Top-1 and Top-5 suggestions). Accuracies are computed over the full denominator of each test set (counting OOV targets as incorrect), and we report exact two-sided paired McNemar $p$-values (with Holm--Bonferroni family-wise correction) alongside $95\%$ Wilson score confidence intervals. Code and scripts are available at \url{https://github.com/leolca/lipspell}.

\begin{table}[ht]
\centering
\small
\caption{Startup candidate-index footprint across all six languages (isolated process measurements).}
\label{tab:index_stats}
\begin{tabular}{lrrrrr}
\toprule
Language & $|V|$ (words) & Build (s) & Delete Keys & Phonetic Keys & Peak RSS (MiB) \\
\midrule
French     &  74,544 &  3.55 &  3.18M & 195k &   783.2 \\
English    &  78,393 &  3.04 &  2.37M & 180k &   497.4 \\
Portuguese & 240,040 & 11.90 &  8.82M & 510k & 1,836.6 \\
Hebrew     & 339,791 &  9.12 &  3.04M & 161k &   971.7 \\
Russian    & 358,624 & 18.46 & 13.92M & 1.27M & 3,456.4 \\
Turkish    & 777,340 & 51.58 & 39.44M & 2.14M & 8,650.6 \\
\bottomrule
\end{tabular}
\end{table}

\section{Results and Discussion}
\label{sec:results}

\subsection{Metric Ablation and Candidate Recall (RQ1)}
Table~\ref{tab:ablation} compares $\ell_1$, salience-weighted $\ell_1$, and cosine across both Phonetic-only mode (ranking by $\lambda_{\text{freq}}\log P(w) - \lambda_{\text{phon}}d_{\text{phon}}$) and full LipSpell. Cosine achieves the highest Phonetic-only Top-1 on all five sets ($+1.11$ to $+3.00$ percentage points over $\ell_1$ and $+1.74$ to $+5.25$\,pp over salience-weighted $\ell_1$) and Phonetic-only Top-5 ($+2.22$ to $+3.33$\,pp over $\ell_1$ and $+1.11$ to $+4.50$\,pp over salience-weighted $\ell_1$) on all five sets, as well as the best LipSpell Top-5 on all five sets (while trailing $\ell_1$/salience on LipSpell Top-1 on Norvig2, $74.75\%$ vs.\ $75.25\%$, and Aspell, $56.40\%$ vs.\ $57.17\%$). Table~\ref{tab:recall} measures candidate-generation recall---the percentage of ground-truth targets $w^*$ retrieved into the scoring pool---comparing orthographic symmetric deletes alone (\emph{SymDelete only}\footnote{Here \emph{SymDelete} denotes Garbe's Symmetric Delete (SymSpell) algorithm \citep{garbe2012symspell}, which retrieves candidates by matching $0$--$2$-character orthographic deletions of the query against pre-indexed deletions of the vocabulary.}), the seven-class phonetic code alone (\emph{Phonetic code only}), and their union (\emph{Combined recall}). The seven-class phonetic code substantially boosts candidate recall beyond SymDelete: $+21.33$ percentage points on Holbrook ($57.00\%\rightarrow 78.33\%$), $+9.11$ pp on Aspell ($79.46\%\rightarrow 88.57\%$), and $+9.60$ pp on PT-Wiki ($86.80\%\rightarrow 96.40\%$, where the phonetic code alone outperforms SymDelete), while evaluating only $204.5$--$398.3$ candidates per query.

\begin{table}[ht]
\centering
\scriptsize
\setlength{\tabcolsep}{3pt}
\caption{Phone-distance metric ablation on English (Phonetic-only Top-1 / Top-5 \% $\mid$ LipSpell Top-1 / Top-5 \%).}
\label{tab:ablation}
\begin{tabular}{lccc}
\toprule
Dataset & $\ell_1$ & Salience-$\ell_1$ & Cosine \\
\midrule
Norvig1 ($n=270$)   & 36.67 / 55.93 $\mid$ 77.41 / 91.85 & 35.93 / 57.04 $\mid$ 77.04 / 92.22 & \textbf{37.78 / 58.15} $\mid$ \textbf{78.52 / 92.59} \\
Norvig2 ($n=400$)   & 33.50 / 62.25 $\mid$ \textbf{75.25} / 88.25 & 31.25 / 61.00 $\mid$ 74.75 / 88.25 & \textbf{36.50 / 65.50} $\mid$ 74.75 / \textbf{88.75} \\
Aspell ($n=516$)    & 24.22 / 43.99 $\mid$ 56.98 / 75.58 & 24.42 / 44.19 $\mid$ \textbf{57.17} / 75.58 & \textbf{26.16 / 46.32} $\mid$ 56.40 / \textbf{75.78} \\
Wikipedia ($n=300$) & 37.67 / 60.00 $\mid$ \textbf{81.67 / 91.33} & 37.67 / 62.00 $\mid$ 81.33 / 91.33 & \textbf{40.67 / 63.33} $\mid$ \textbf{81.67 / 91.33} \\
Holbrook ($n=300$)  & 11.33 / 23.00 $\mid$ 30.00 / 45.33 & 11.00 / 22.67 $\mid$ 30.67 / 45.67 & \textbf{13.00 / 25.33} $\mid$ \textbf{32.00 / 46.67} \\
\bottomrule
\end{tabular}
\end{table}

\begin{table}[ht]
\centering
\small
\caption{Two-stage candidate-generation recall (\%) and average candidates per query.}
\label{tab:recall}
\begin{tabular}{lccccccc}
\toprule
Metric & Norvig1 & Norvig2 & Aspell & Wiki-EN & Holbrook & TOEFL & PT-Wiki \\
\midrule
SymDelete only (\%)       & 96.67 & 89.25 & 79.46 & 93.33 & 57.00 & 93.40 & 86.80 \\
Phonetic code only (\%)  & 71.48 & 69.75 & 62.40 & 73.00 & 56.33 & 61.30 & 89.20 \\
Combined recall (\%)      & \textbf{98.52} & \textbf{95.00} & \textbf{88.57} & \textbf{94.67} & \textbf{78.33} & \textbf{94.90} & \textbf{96.40} \\
Avg.\ candidates / query  & 262.9 & 219.0 & 235.5 & 204.5 & 398.3 & 239.8 & 280.5 \\
\bottomrule
\end{tabular}
\end{table}

\subsection{Component Ablation (RQ2)}
Table~\ref{tab:comp_ablation} isolates each signal. Adding phonetics (\textbf{A}$\rightarrow$\textbf{C}) improves Top-1 substantially on Norvig1 ($+4.45$ pp), Aspell ($+3.30$ pp), Holbrook ($+4.33$ pp), and especially PT-Wiki ($+12.80$ pp). Combining all signals (\textbf{E}) achieves the highest Top-1 and Top-5 on Norvig1, Holbrook, and PT-Wiki (with NFD tolerance; without NFD tolerance, \textbf{B} reaches $63.80\% / 81.80\%$ and \textbf{E} reaches $73.20\% / 85.20\%$), and highest Top-1 on Aspell and TOEFL, while removing frequency (\textbf{D}) degrades Top-1 by $3.88$--$8.00$ pp on English sets.

\begin{table}[ht]
\centering
\scriptsize
\setlength{\tabcolsep}{3.5pt}
\caption{Component ablation (Top-1 \% / Top-5 \%). PT-Wiki values include NFD diacritic tolerance.}
\label{tab:comp_ablation}
\begin{tabular}{lccccc}
\toprule
Configuration & Norvig1 & Aspell & Holbrook & TOEFL & PT-Wiki \\
\midrule
\textbf{A}: Orth + Freq              & 71.11 / 89.26 & 52.71 / 72.67 & 26.67 / 42.33 & 76.10 / 91.00 & 61.00 / 81.40 \\
\textbf{B}: Orth + Kb + Freq         & 74.81 / 91.48 & 53.49 / 74.42 & 27.00 / 44.33 & 76.80 / \textbf{91.50} & 68.40 / 83.00 \\
\textbf{C}: Orth + Phon + Freq       & 75.56 / \textbf{92.59} & 56.01 / 75.39 & 31.00 / 45.00 & 76.10 / 91.00 & 73.80 / 85.00 \\
\textbf{D}: Orth + Kb + Phon         & 73.33 / 90.00 & 52.52 / 73.84 & 25.33 / 41.33 & 69.30 / 87.60 & \textbf{76.20} / 84.80 \\
\textbf{E}: Full LipSpell            & \textbf{78.52 / 92.59} & \textbf{56.40 / 75.78} & \textbf{32.00 / 46.67} & \textbf{77.30} / 91.10 & \textbf{76.20 / 85.20} \\
\bottomrule
\end{tabular}
\end{table}

\subsection{Comparison with Aspell, Hunspell, and Norvig (RQ2, RQ3)}
Tables~\ref{tab:english_cmp} and~\ref{tab:ptbr} compare LipSpell against Norvig, Aspell, and Hunspell, including target vocabulary coverage for each system (\texttt{Ours/Asp/Hun}). On English, LipSpell outperforms Norvig and Hunspell on every set (paired McNemar $p < 10^{-4}$ against Hunspell on five of six sets, surviving Holm--Bonferroni correction \citep{holm1979simple} across all pairwise comparisons) and beats Aspell on Norvig1 ($78.52\%$ vs.\ $75.56\%$), Holbrook ($32.00\%$ vs.\ $26.67\%$, $p=0.0226$), and TOEFL-Spell ($77.30\%$ vs.\ $72.80\%$, $p=0.0011$), while trailing Aspell Top-1 on Norvig2 ($74.75\%$ vs.\ $77.25\%$), the Aspell benchmark ($56.40\%$ vs.\ $58.72\%$), and Wikipedia-EN ($81.67\%$ vs.\ $83.33\%$). On Wikipedia-EN, Aspell's dictionary covers only $70.3\%$ of targets compared to $97.3\%$ for our lexicon, yet Aspell achieves slightly higher Top-1 precision on the covered subset, whereas LipSpell achieves the highest Top-5 recall ($91.33\%$ vs.\ $91.00\%$).

On Brazilian Portuguese, even though Aspell and Hunspell have higher target vocabulary coverage ($98.4\%$ vs.\ $96.8\%$ on $n=500$; $97.4\%$ vs.\ $95.8\%$ on the full $n=1{,}149$ A--Z set), LipSpell (with fixed English weights) achieves $73.20\pm3.88\%$ Top-1 ($85.20\%$ Top-5) on $n=500$ ($p \le 4.53\times 10^{-5}$ vs.\ all three baselines) and $72.24\pm2.59\%$ Top-1 on the full $n=1{,}149$ A--Z set vs.\ $66.23\%$ for Norvig ($p=2.08\times 10^{-13}$), $63.19\%$ for Hunspell ($p=1.46\times 10^{-9}$), and $60.14\%$ for Aspell ($p=4.21\times 10^{-16}$). Note that these full-slice figures include OOV targets and uncleaned tokens, whereas Table~\ref{tab:multilingual} evaluates clean in-vocabulary $50/50$ non-word splits with per-language tuning.

To isolate where LipSpell's phonetic and keyboard signals deliver the largest improvements over purely orthographic ranking, Table~\ref{tab:multi_edit} stratifies Top-1 accuracy across all six English benchmarks by Damerau--Levenshtein edit distance ($d_{\text{orth}} \le 1$ vs.\ $d_{\text{orth}} \ge 2$). On single-edit errors ($d_{\text{orth}} \le 1$), LipSpell improves over Norvig by $+0.49$ to $+3.06$\,pp on five of six English datasets (and trails slightly by $-0.76$\,pp on TOEFL-Spell). On multi-edit errors ($d_{\text{orth}} \ge 2$), however, LipSpell delivers consistent, substantial gains on every single benchmark: $+12.12$ pp on Norvig1 ($65.15\%$ vs.\ $53.03\%$), $+10.20$ pp on Wikipedia-EN ($57.14\%$ vs.\ $46.94\%$), $+6.75$ pp on Aspell ($35.86\%$ vs.\ $29.11\%$), $+5.84$ pp on Holbrook ($13.14\%$ vs.\ $7.30\%$), $+5.56$ pp on Norvig2 ($48.89\%$ vs.\ $43.33\%$), $+3.77$ pp on TOEFL-Spell ($42.92\%$ vs.\ $39.15\%$), and $+15.88$ pp on full PT-Wiki ($43.92\%$ vs.\ $28.04\%$, $n=189$). These stratified results confirm that distinctive-feature phonetic scoring is especially effective precisely where character-level edit distance loses discriminative power.

\begin{table}[ht]
\centering
\scriptsize
\setlength{\tabcolsep}{3.5pt}
\caption{Stratified Top-1 accuracy (\%) by orthographic edit distance ($d_{\text{orth}} \le 1$ vs.\ $d_{\text{orth}} \ge 2$) across English benchmarks.}
\label{tab:multi_edit}
\begin{tabular}{lccccccc}
\toprule
Dataset & $n_{d\le 1}$ & Norvig ($d\le 1$) & LipSpell ($d\le 1$) & $n_{d\ge 2}$ & Norvig ($d\ge 2$) & LipSpell ($d\ge 2$) & Gain ($d\ge 2$) \\
\midrule
Norvig1     & 204 & 82.35 & \textbf{82.84} &  66 & 53.03 & \textbf{65.15} & $+12.12$ pp \\
Norvig2     & 310 & 79.68 & \textbf{82.26} &  90 & 43.33 & \textbf{48.89} & $+5.56$ pp \\
Aspell      & 279 & 72.76 & \textbf{73.84} & 237 & 29.11 & \textbf{35.86} & $+6.75$ pp \\
Wikipedia   & 251 & 84.06 & \textbf{86.45} &  49 & 46.94 & \textbf{57.14} & $+10.20$ pp \\
Holbrook    & 163 & 44.79 & \textbf{47.85} & 137 &  7.30 & \textbf{13.14} & $+5.84$ pp \\
TOEFL-Spell & 788 & \textbf{87.31} & 86.55 & 212 & 39.15 & \textbf{42.92} & $+3.77$ pp \\
\bottomrule
\end{tabular}
\end{table}

Table~\ref{tab:pt_error_taxonomy} provides a complementary hierarchical error-category breakdown on the complete $n=1{,}149$ Brazilian Portuguese Wikipedia dataset, comparing LipSpell (with fixed English weights and NFD diacritic tolerance) against Norvig (which achieved the strongest baseline score and shares the exact same $95.8\%$ lexicon coverage). More than half of the Portuguese errors ($54.66\%$) are purely diacritical (\textit{acessivel} $\rightarrow$ \textit{acess\'ivel}), where LipSpell outperforms Norvig by $+11.30$ pp; on non-diacritic multi-edit errors ($d_{\text{orth}} \ge 2$, $n=163$), LipSpell achieves $+15.34$ pp over Norvig ($42.33\%$ vs.\ $26.99\%$).

\begin{table}[ht]
\centering
\small
\caption{Brazilian Portuguese error taxonomy on full PT-Wiki ($n=1{,}149$) and Top-1 accuracy (\%) against Norvig.}
\label{tab:pt_error_taxonomy}
\begin{tabular}{lrrccc}
\toprule
Error Category & $n$ & Share (\%) & Norvig (\%) & LipSpell (\%) & Gain (pp) \\
\midrule
Diacritic-only ($d_{\text{NFD}}=0$) & 628 & 54.66 & 67.04 & \textbf{78.34} & $+11.30$ \\
Substitution ($d=1$)              & 222 & 19.32 & 81.08 & \textbf{82.43} & $+1.35$ \\
Ins./Del./Transp. ($d=1$)         & 136 & 11.84 & \textbf{85.29} & 84.56 & $-0.73$ \\
Multi-edit ($d\ge 2$)             & 163 & 14.19 & 26.99 & \textbf{42.33} & $+15.34$ \\
\bottomrule
\end{tabular}
\end{table}

\begin{table}[ht]
\centering
\small
\setlength{\tabcolsep}{4.5pt}
\caption{English comparison on the full benchmarks (Top-1 \% / Top-5 \%) and target vocabulary coverage (\texttt{Ours/Asp/Hun} \%).}
\label{tab:english_cmp}
\begin{tabular}{lccccc}
\toprule
Dataset & Vocab (\texttt{O/A/H}\%) & Aspell & Hunspell & Norvig & LipSpell \\
\midrule
Norvig1 ($n=270$)      & 98.5 / 98.5 / 99.3 & 75.56 / \textbf{92.59} & 64.44 / 84.07 & 75.19 / 91.11 & \textbf{78.52 / 92.59} \\
Norvig2 ($n=400$)      & 97.0 / 95.5 / 96.0 & \textbf{77.25} / 87.00 & 65.50 / 84.00 & 71.50 / 87.50 & 74.75 / \textbf{88.50} \\
Aspell ($n=516$)       & 99.0 / 97.7 / 98.1 & \textbf{58.72 / 80.81} & 53.49 / 73.64 & 52.71 / 71.12 & 56.40 / 75.78 \\
Wikipedia ($n=300$)    & 97.3 / 70.3 / 71.7 & \textbf{83.33} / 91.00 & 55.33 / 70.00 & 78.00 / 90.33 & 81.67 / \textbf{91.33} \\
Holbrook ($n=300$)     & 97.0 / 86.3 / 86.3 & 26.67 / 42.33 & 17.33 / 31.67 & 27.67 / 41.67 & \textbf{32.00 / 46.67} \\
TOEFL-Spell ($n=1000$) & 98.7 / 97.5 / 97.8 & 72.80 / 90.40 & 66.90 / 86.90 & 77.10 / 90.40 & \textbf{77.30 / 91.10} \\
\bottomrule
\end{tabular}
\end{table}

\begin{table}[ht]
\centering
\scriptsize
\setlength{\tabcolsep}{3pt}
\caption{Brazilian Portuguese evaluation on Wikipedia typos ($n=500$ slice and full $n=1{,}149$ A--Z set).}
\label{tab:ptbr}
\begin{tabular}{lccccc}
\toprule
System & Coverage ($500$/$1149$) & Top-1 ($n=500$) & Top-5 ($n=500$) & Top-1 ($n=1149$) & Top-5 ($n=1149$) \\
\midrule
Aspell (\texttt{pt\_BR})            & 98.4\% / 97.4\% & 64.00\% & 76.80\% & 60.14\% & 72.93\% \\
Hunspell (\texttt{pt\_BR})          & 98.4\% / 97.4\% & 64.20\% & 79.00\% & 63.19\% & 76.33\% \\
Norvig                              & 96.8\% / 95.8\% & 66.40\% & 83.00\% & 66.23\% & 81.29\% \\
Phonetic-only                       & 96.8\% / 95.8\% & 39.20\% & 68.20\% & 39.16\% & 68.67\% \\
\textbf{LipSpell} (direct transfer) & 96.8\% / 95.8\% & 73.20\% & \textbf{85.20\%} & 72.24\% & 83.72\% \\
\textbf{LipSpell (NFD)}             & 96.8\% / 95.8\% & \textbf{76.20\%} & \textbf{85.20\%} & \textbf{74.93\%} & \textbf{83.90\%} \\
\bottomrule
\end{tabular}
\end{table}

\subsection{Multilingual Pilots (Slavic, Semitic, Turkic, Romance) and Per-Language Tuning}
To stress-test portability across distinct language families and writing systems, we pair the language-specific frontends and lexicons from Sections~3.1 and~4 with targeted phonological and orthographic refinements: word-final obstruent devoicing in \texttt{RussianG2P}, grave-vowel normalization (\texttt{è/ì/ò/ù}) and word-final \texttt{-z} devoicing in \texttt{PortugueseG2P}, a $0.25$ indel cost for Hebrew \textit{matres lectionis} (\textit{alef}/\textit{vav}/\textit{yod}), nasal-vowel rules in \texttt{FrenchG2P}, and universal Unicode NFD diacritic-drop tolerance ($0.15$ cost) in \texttt{KeyboardLayout} (summarized in Table~\ref{tab:lang_components}).

\begin{table}[ht]
\centering
\scriptsize
\caption{Language-specific resources and refinements paired with the common LipSpell scoring engine.}
\label{tab:lang_components}
\begin{tabular}{lcccccc}
\toprule
Component & English & Portuguese & Russian & Hebrew & Turkish & French \\
\midrule
Rule-based G2P & \checkmark & \checkmark & \checkmark & \checkmark & \checkmark & \checkmark \\
Phonological / orthographic rules & \checkmark & \checkmark & \checkmark & --- & --- & \checkmark \\
Weak-character indel overrides & --- & --- & --- & \checkmark & --- & --- \\
Pronunciation dictionary & CMUdict & --- & --- & --- & --- & \texttt{ipa-dict} \\
\bottomrule
\end{tabular}
\end{table}

\begin{figure}[ht]
\centering
\includegraphics[width=\textwidth]{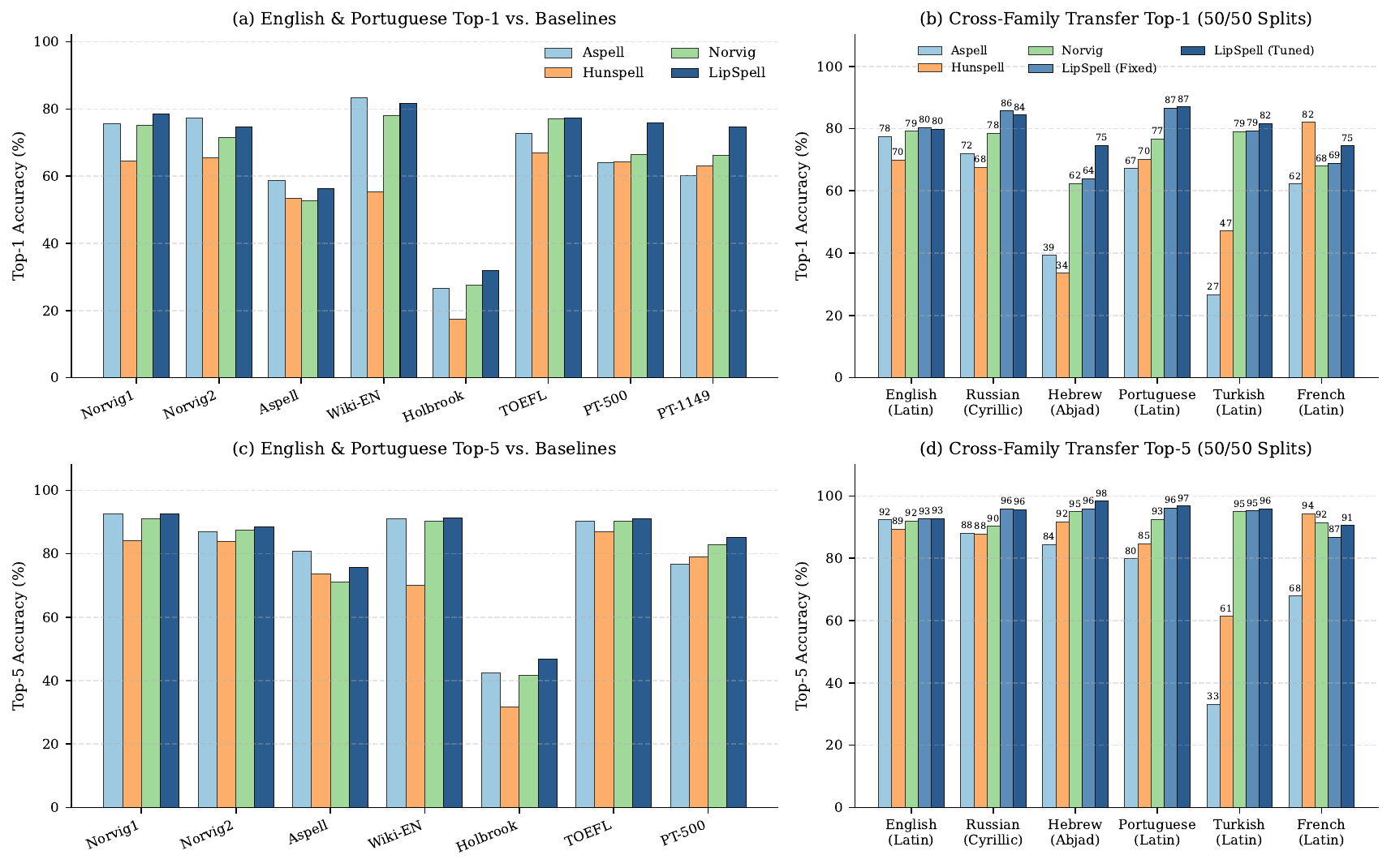}
\caption{Summary of Top-1 (a, b) and Top-5 (c, d) accuracy across English and Brazilian Portuguese benchmarks (a, c) and cross-family transfer on 50/50 clean non-word splits (b, d), comparing \textbf{LipSpell} (Cosine, fixed English weights and per-language tuned weights) against Aspell, Hunspell, and Norvig.}
\label{fig:overview}
\end{figure}

Table~\ref{tab:multilingual} (visualized alongside the English and Portuguese full-benchmark baselines in Figure~\ref{fig:overview}) compares Aspell, Hunspell, Norvig, LipSpell (Fixed), and LipSpell (Tuned) on 50/50 clean non-word splits, where \emph{LipSpell (Fixed)} applies the default English-tuned channel weights $(1.0, 4.0, 1.5, 4.0)$ across languages without per-language adjustment, whereas \emph{LipSpell (Tuned)} selects the channel weights $\boldsymbol{\lambda}$ (rightmost column) via grid search on each language's held-out $50\%$ development split before evaluating on the disjoint $50\%$ test split. Clean lexicons, per-language G2P refinements, and weak-character indel overrides substantially boost performance across all four languages without degrading English ($78.52\%$ vs.\ $75.19\%$ Norvig on Norvig1). LipSpell (Tuned) outperforms Aspell on all six languages ($+2.24$ to $+55.00$ percentage points Top-1 from the Tuned column, where official \texttt{aspell-tr} covers only $80{,}261$ uninflected forms) and outperforms Hunspell on five of the six languages by $+9.81$ to $+40.98$ percentage points Top-1 from the Tuned column (excluding French at $-7.55$\,pp): English (with CMUdict) reaches $80.37\%$ fixed ($79.75\%$ tuned vs.\ $77.51\%$ Aspell / $69.94\%$ Hunspell), Russian reaches $85.67\%$ with Fixed English weights (vs.\ $71.96\%$ Aspell / $67.60\%$ Hunspell, $p=1.52\times 10^{-5}$ vs.\ Norvig; per-language tuning on the small dev split slightly overfits to $84.42\%$), Hebrew jumps to $74.59\%$ tuned (vs.\ $39.34\%$ Aspell / $33.61\%$ Hunspell, $98.36\%$ Top-5), Portuguese reaches $86.97\%$ tuned (vs.\ $67.13\%$ Aspell / $70.14\%$ Hunspell), and Turkish reaches $81.67\%$ tuned (vs.\ $26.67\%$ Aspell / $47.14\%$ Hunspell), while Hunspell leads on French ($82.08\%$ vs.\ $74.53\%$ tuned LipSpell with \texttt{ipa-dict} and $62.26\%$ Aspell) due to its rich morphological affix rules. Notably, the tuned $\boldsymbol{\lambda}$ weights reflect orthographic depth: opaque orthographies (Portuguese and French) emphasize phonetics ($\lambda_{\text{phon}}=6$), transparent Turkish emphasizes orthography ($\lambda_{\text{orth}}=8, \lambda_{\text{kb}}=4$), and consonantal Hebrew balances keyboard proximity and phonetics ($\lambda_{\text{kb}}=4, \lambda_{\text{phon}}=1$). Furthermore, Aspell and Hunspell both achieve high target coverage on Hebrew ($93.44\%$ and $100.00\%$) and strong Top-5 recall ($84.43\%$ and $91.80\%$), so their low Hebrew Top-1 ($39.34\%$ and $33.61\%$) stems from ranking ties among unvocalized consonant-root neighbors without frequency priors or \textit{matres lectionis} weighting; on Turkish, \texttt{aspell-tr} accepts only $32.14\%$ of inflected targets ($80{,}261$ static forms), whereas Hunspell accepts $89.76\%$ via affixes yet achieves only $47.14\%$ Top-1 / $61.43\%$ Top-5 without corpus frequency priors.

\begin{table}[ht]
\centering
\scriptsize
\setlength{\tabcolsep}{2.2pt}
\caption{Cross-lingual transfer across three language families and three scripts using \textbf{LipSpell} (Cosine metric, 50/50 clean in-vocabulary non-word splits, including TOEFL-Spell $n=489$; Top-1 / Top-5 \%).}
\label{tab:multilingual}
\begin{tabular}{lccccccc}
\toprule
Language (Family / Script) & $n$ & Aspell & Hunspell & Norvig & LipSpell (Fixed) & LipSpell (Tuned) & $\boldsymbol{\lambda}$ \\
\midrule
English (Germanic / Latin)   & 489 & 77.51 / 92.43 & 69.94 / 89.37 & 79.14 / 92.02 & \textbf{80.37} / 92.64 & 79.75 / \textbf{92.84} & $(1,4,2,2)$ \\
Russian (Slavic / Cyrillic)  & 321 & 71.96 / 88.16 & 67.60 / 87.85 & 78.50 / 90.34 & \textbf{85.67 / 95.95} & 84.42 / 95.64 & $(1,2,4,1)$ \\
Hebrew (Semitic / Abjad)     & 122 & 39.34 / 84.43 & 33.61 / 91.80 & 62.30 / 95.08 & 63.93 / 95.90 & \textbf{74.59 / 98.36} & $(1,4,4,1)$ \\
Portuguese (Romance / Latin) & 499 & 67.13 / 79.96 & 70.14 / 84.77 & 76.75 / 92.59 & 86.77 / 96.19 & \textbf{87.17 / 96.79} & $(1,4,1,6)$ \\
Turkish (Turkic / Latin)     & 420 & 26.67 / 33.10 & 47.14 / 61.43 & 79.05 / 95.00 & 79.29 / 95.24 & \textbf{81.67 / 95.95} & $(1,8,4,1)$ \\
French (Romance / Latin)     & 106 & 62.26 / 67.92 & \textbf{82.08 / 94.34} & 67.92 / 91.51 & 68.87 / 86.79 & 74.53 / 90.57 & $(1,4,4,6)$ \\
\bottomrule
\end{tabular}
\end{table}

Table~\ref{tab:multilingual_cov} reports target-vocabulary coverage for Aspell and Hunspell alongside LipSpell's development-split Top-1 (Fixed / Tuned) and $95\%$ Wilson confidence intervals on the best test-split Top-1 (LipSpell and Norvig have $100\%$ target coverage on these splits by construction). Tuned gains generalize well from dev to test on Portuguese ($88.96\% \rightarrow 87.17\%$), Turkish ($78.10\% \rightarrow 81.67\%$), Hebrew ($79.34\% \rightarrow 74.59\%$), and French ($76.42\% \rightarrow 74.53\%$), whereas wider Wilson intervals on Hebrew ($n=122$) and French ($n=106$) reflect their smaller test samples.

\begin{table}[ht]
\centering
\scriptsize
\caption{Target-vocabulary coverage (Aspell / Hunspell \%), dev-split Top-1 (Fixed / Tuned \%), and $95\%$ Wilson CI on best LipSpell test Top-1.}
\label{tab:multilingual_cov}
\begin{tabular}{lccccc}
\toprule
Language & $n_{\text{dev}} / n_{\text{test}}$ & Aspell Cov (\%) & Hunspell Cov (\%) & Dev Top-1 (Fix / Tun) & Best Test Top-1 (95\% CI) \\
\midrule
English    & 489 / 489 & 98.77 & 98.77 & 77.71 / 79.55 & 80.37 [76.62, 83.65] \\
Russian    & 320 / 321 & 98.75 & 99.07 & 85.62 / 85.62 & 85.67 [81.41, 89.08] \\
Hebrew     & 121 / 122 & 93.44 & 100.00 & 70.25 / 79.34 & 74.59 [66.20, 81.48] \\
Portuguese & 498 / 499 & 99.40 & 99.40 & 85.94 / 88.96 & 87.17 [83.95, 89.83] \\
Turkish    & 420 / 420 & 32.14 & 89.76 & 75.48 / 78.10 & 81.67 [77.69, 85.08] \\
French     & 106 / 106 & 76.42 & 95.28 & 68.87 / 76.42 & 74.53 [65.48, 81.87] \\
\bottomrule
\end{tabular}
\end{table}

\subsection{Rule-Based G2P vs.\ External Pronunciation Dictionaries}
To test whether external IPA pronunciation dictionaries improve over purely rule-based G2P, we evaluated Rule-only G2P against seven Dictionary + Rule fallback configurations across the six languages (\texttt{CMUdict} and \texttt{ipa-dict} \citep{opendictipadict} for English; \texttt{ipa-dict} for Portuguese and French; \texttt{WikiPron} \citep{lee2020wikipron} for Russian, Hebrew, and Turkish) on the same 50/50 splits (Table~\ref{tab:g2p_ablation}). Before lookup, external dictionary entries undergo canonical Unicode NFD normalization to strip combining diacritics, primary/secondary stress marks (/\textprimstress, \textsecstress/), and length markers (/\textlengthmark/), and narrow allophones outside the $45$-segment feature inventory (such as /\textinvscr/$\rightarrow$/\textturnr/, /r/ and /\textipa{;R}/$\rightarrow$/\textfishhookr/, /\textltilde/$\rightarrow$/l/, /\textgamma/$\rightarrow$/\textipa{g}/) are remapped to their closest inventory segments; any word absent from the external dictionary falls back to the language's rule-based G2P.

Across the non-French languages, rule-only G2P matches or outperforms external dictionaries in most configurations (e.g., Russian $85.67\%$ vs.\ $83.80\%$ fixed; Portuguese $87.17\%$ vs.\ $85.37\%$ tuned; on English, CMUdict improves Fixed Top-1 from $79.35\%$ to $80.37\%$ while Rule-only edges out CMUdict on Tuned Top-1, $79.96\%$ vs.\ $79.75\%$). This pattern arises because passing both the misspelled query and lexicon candidates through the exact same deterministic G2P rules cancels systematic rule approximations and avoids cross-convention mismatches between narrow dictionary transcriptions and the rule-based OOV fallback. French is the sole exception where an opaque orthography benefits consistently from \texttt{ipa-dict} ($245\text{k}$ entries), improving tuned accuracy from $73.58\% / 89.62\%$ to $74.53\% / 90.57\%$ ($+0.95$\,pp Top-1 and Top-5; $+1.88$\,pp fixed Top-5) because silent letters and irregular vowel digraphs cannot be predicted reliably by compact surface rules alone. Thus, LipSpell operates cleanly with zero external pronunciation dictionaries or selectively integrates one for opaque orthographies.

\begin{table}[ht]
\centering
\scriptsize
\setlength{\tabcolsep}{2.8pt}
\caption{Rule-only G2P vs.\ Pronunciation Dictionary + Rule fallback across seven configurations on 50/50 clean non-word splits (Top-1 / Top-5 \%).}
\label{tab:g2p_ablation}
\begin{tabular}{lrcccccc}
\toprule
Language & $n$ & Dictionary & Rule-only Fixed & Dict+Rule Fixed & Rule-only Tuned & Dict+Rule Tuned \\
\midrule
English    & 489 & CMUdict (\texttt{0.7b})  & 79.35 / \textbf{93.25} & \textbf{80.37} / 92.64 & \textbf{79.96 / 93.46} & 79.75 / 92.84 \\
English    & 489 & \texttt{ipa-dict} (126k) & 79.35 / \textbf{93.25} & \textbf{79.96} / 92.02 & \textbf{79.96 / 93.46} & 79.75 / 92.02 \\
Russian    & 321 & WikiPron (467k)   & \textbf{85.67 / 95.95} & 83.80 / 95.02 & \textbf{84.42 / 95.64} & 83.80 / 95.02 \\
Hebrew     & 122 & WikiPron (6.8k)   & \textbf{63.93 / 95.90} & \textbf{63.93} / 95.08 & \textbf{74.59 / 98.36} & \textbf{74.59 / 98.36} \\
Portuguese & 499 & \texttt{ipa-dict} (96k)  & \textbf{86.77 / 96.19} & 85.77 / 95.39 & \textbf{87.17 / 96.79} & 85.37 / 95.99 \\
Turkish    & 420 & WikiPron (12.3k)  & \textbf{79.29 / 95.24} & \textbf{79.29 / 95.24} & \textbf{81.67 / 95.95} & \textbf{81.67 / 95.95} \\
French     & 106 & \texttt{ipa-dict} (245k) & 67.92 / 84.91 & \textbf{68.87 / 86.79} & 73.58 / 89.62 & \textbf{74.53 / 90.57} \\
\bottomrule
\end{tabular}
\end{table}

\subsection{Qualitative Case Studies and Error Analysis}
Table~\ref{tab:case_studies} illustrates representative Top-1 predictions from Norvig and LipSpell on real English and Brazilian Portuguese misspellings. On multi-edit cognitive errors, purely orthographic ranking prefers high-frequency $d_{\text{orth}}=1$ or closer $d_{\text{orth}}=2$ competitors: for English \textit{semetary} $\rightarrow$ \textit{cemetery} ($d_{\text{orth}}=2$), Norvig selects \textit{secretary}, whereas LipSpell recognizes that \texttt{c}+\texttt{e} and \texttt{s} share /\text{s}/ and recovers \textit{cemetery}; similarly, for \textit{proplen} $\rightarrow$ \textit{problem}, voicing/nasal proximity between /p/ and /b/ and between /n/ and /m/ overrides Norvig's higher-frequency neighbor \textit{people}. In Portuguese, \textit{albun} $\rightarrow$ \textit{\'album} ($d_{\text{orth}}=2$: \textit{u}/\textit{\'u} and \textit{n}/\textit{m}) is miscorrected by Norvig to \textit{algum} (also at $d_{\text{orth}}=2$: \textit{b}/\textit{g} and \textit{n}/\textit{m}, but with higher corpus frequency), while LipSpell combines nasal proximity (/n/ $\leftrightarrow$ /m/) with NFD diacritic tolerance (\textit{u}/\textit{\'u}) to rank \textit{\'album} first; likewise, \textit{aterisagem} $\rightarrow$ \textit{aterrissagem} ($d_{\text{orth}}=2$) succeeds because \texttt{rr} $\rightarrow$ /h/ and \texttt{ss} $\rightarrow$ /s/ neutralize doubled-consonant errors phonetically.

Conversely, the bottom two rows of Table~\ref{tab:case_studies} highlight characteristic failure modes of the current isolated-word formulation. In Portuguese \textit{beiges} $\rightarrow$ \textit{beges}, LipSpell promotes the higher-frequency word \textit{beijos} because the phonetic distance between /\text{be\textyogh is}/ and /\text{bej\textyogh us}/ is small enough for $\log P(w)$ to tip the balance over $d_{\text{orth}}=1$. In \textit{dijitara} $\rightarrow$ \textit{digitara}, both forms share the exact same phonetic sequence (/di\textyogh ita\textfishhookr a/), and the much higher corpus frequency of the infinitive \textit{digitar} outweighs the single-character deletion cost relative to the inflected pluperfect \textit{digitara}.

\begin{table}[ht]
\centering
\scriptsize
\setlength{\tabcolsep}{3pt}
\caption{Qualitative Top-1 comparison on real English and Brazilian Portuguese misspellings.}
\label{tab:case_studies}
\begin{tabular}{llllcl}
\toprule
Language & Input $x$ & Target $w^*$ & Norvig Top-1 & LipSpell Top-1 & Primary Mechanism \\
\midrule
English    & \textit{proplen}    & \textit{problem}      & \textit{people}    & \textbf{\textit{problem}}      & Place/voicing (/p/$\leftrightarrow$/b/, /n/$\leftrightarrow$/m/) \\
English    & \textit{semetary}   & \textit{cemetery}     & \textit{secretary} & \textbf{\textit{cemetery}}     & Contextual sibilant (\texttt{c+e}$\rightarrow$/s/) \\
English    & \textit{splened}    & \textit{splendid}     & \textit{opened}    & \textbf{\textit{splendid}}     & Tense/lax vowel (/\textepsilon/$\leftrightarrow$/\textsci/) \\
English    & \textit{awfall}     & \textit{awful}        & \textit{wall}      & \textbf{\textit{awful}}        & Vowel/liquid alignment ($d_{\text{orth}}=2$) \\
Portuguese & \textit{albun}      & \textit{\'album}      & \textit{algum}     & \textbf{\textit{\'album}}      & Nasal /n/$\leftrightarrow$/m/ + NFD tolerance \\
Portuguese & \textit{aterisagem} & \textit{aterrissagem} & \textit{aterragem} & \textbf{\textit{aterrissagem}} & Digraph equivalence (\texttt{rr, ss}) \\
\midrule
Portuguese & \textit{beiges}     & \textit{beges}        & \textbf{\textit{beges}}     & \textit{beijos}                & Loss: high freq.\ of phonetically close neighbor \\
Portuguese & \textit{dijitara}   & \textit{digitara}     & \textbf{\textit{digitara}}  & \textit{digitar}               & Loss: homophonous infinitive frequency bias \\
\bottomrule
\end{tabular}
\end{table}

\paragraph{Implications for L2 Learner Error Correction.}
The TOEFL-Spell results ($77.30\%$ Top-1 vs.\ $72.80\%$ Aspell, $p=0.0011$; $+3.77$\,pp over Norvig on multi-edit errors $d_{\text{orth}} \ge 2$, where LipSpell improves over Norvig on $13$ of the $212$ pairs and regresses on $5$, yielding $+8$ net corrections; and $80.37\%$ fixed / $92.84\%$ tuned Top-5 on the clean $50/50$ non-word split) demonstrate that distinctive-feature phonetic scoring is directly effective for non-native (L2) learner writing, where writers' errors frequently reflect phonetic approximations rather than keyboard slips. Moreover, because substitution costs are derived explicitly from the $|P|\times|P|$ IPA distinctive-feature matrix $M$, the framework naturally supports L1-transfer adaptation: systematic learner confusions rooted in a writer's native phonology (such as /l/$\leftrightarrow$/\textturnr/, /p/$\leftrightarrow$/b/, /v/$\leftrightarrow$/b/, or tense/lax vowel mergers) can be accommodated by adjusting feature salience weights $\mathbf{w}$ or targeted entries in $M$ without retraining a character-level noisy channel from scratch.

\subsection{Limitations and Practical Deployment Considerations}
\label{sec:limitations}
Five practical and methodological limitations remain:
\begin{enumerate}
    \item \textbf{Isolated-word scope:} Like any single-word spell checker, scoring without surrounding sentence context cannot resolve real-word errors (e.g., \textit{their}/\textit{there} in English or \textit{mas}/\textit{mais} in Portuguese), nor can it disambiguate homophonous inflections when corpus priors dominate.
    \item \textbf{Distorted G2P inputs:} Severely mangled inputs or non-standard consonant clusters can trigger misleading rule-based G2P transcriptions, occasionally degrading phonetic matching relative to pure orthographic ranking.
    \item \textbf{One-to-one phone alignment:} Standard DP aligns one IPA segment to one IPA segment, preventing direct $1$-to-$2$ merges for diphthongs, affricated clusters, or vocalized glides without explicit multi-segment DP transitions.
    \item \textbf{Single pronunciation per word:} Assigning one canonical sequence $\phi(w)$ rather than a pronunciation set $\Phi(w)$ leaves dialectal and lexical variants (such as Portuguese \texttt{x} in \textit{t\'axi} /ks/ vs.\ \textit{exame} /z/) unmodeled unless an external lexicon is used.
    \item \textbf{Memory footprint on agglutinative lexicons:} While English ($497.4$\,MiB) and French ($783.2$\,MiB) fit comfortably in lightweight environments, indexing $0$--$2$-character deletions over $777\text{k}$ inflected Turkish forms requires $8.45$\,GiB ($8.65$\,GB) peak RSS; production deployments on agglutinative languages may therefore restrict deletion depth to $1$ for long words or adopt compressed trie/DAWG storage.
\end{enumerate}

\section{Conclusion}
\label{sec:conclusion}
We presented a language-independent phonetic scoring framework combining precomputed IPA distinctive-feature edit distances, keyboard geometry, and corpus frequency priors. Component ablations confirm that keyboard and phonetic signals provide complementary improvements over an orthographic-frequency baseline: on PT-Wiki, NFD diacritic tolerance contributes substantially (Full LipSpell drops from $76.20\%$ to $73.20\%$ Top-1 when NFD tolerance is disabled), whereas keyboard geometry on its own adds $+2.40$\,pp ($73.80\% \rightarrow 76.20\%$). Cosine distance over ternary distinctive features achieves the strongest phonetic-only Top-1/Top-5 and unified Top-5 accuracy across English benchmarks ($78.52\%$ on Norvig1, $77.30\%$ on TOEFL-Spell). By swapping only language-specific G2P, keyboard, and lexicon resources, the same scoring engine achieves strong fixed-weight transfer across six languages spanning three top-level language families (Indo-European, Afro-Asiatic, and Turkic; five branches: Germanic, Romance, Slavic, Semitic, and Turkic) and three writing systems---outperforming Norvig, Aspell, and Hunspell on Brazilian Portuguese ($73.20\%$ without NFD tolerance and $76.20\%$ with NFD tolerance vs.\ $66.40\%$/$64.00\%$/$64.20\%$ on $n=500$), beating Aspell on all six languages and Hunspell on five of six on 50/50 non-word splits, and surpassing Norvig on Russian Cyrillic ($85.67\%$ fixed vs.\ $78.50\%$, $p=1.52\times 10^{-5}$), Portuguese ($87.17\%$ vs.\ $76.75\%$), Turkish ($81.67\%$ vs.\ $79.05\%$), Hebrew Abjad ($74.59\%$ vs.\ $62.30\%$), and French ($74.53\%$ vs.\ $67.92\%$). Per-language weight tuning adapts the channel balance to orthographic depth, while G2P ablations show that purely rule-based G2P matches or outperforms external dictionaries across the non-French languages (with CMUdict helping English Fixed Top-1) and \texttt{ipa-dict} further improves opaque French; large agglutinative vocabularies such as Turkish ($8.45$\,GiB / $8.65$\,GB peak RSS) remain the main practical memory cost. Future work will integrate Epitran \citep{mortensen2018epitran}, $1$-to-$2$ phone alignments, lightweight $n$-gram context for real-word errors, and L1-specific feature-weight adaptation for L2 learner spelling correction.

\bibliographystyle{plainnat}
\bibliography{refs}

\appendix

\section{Complete Rule-Based G2P Specifications}
\label{app:g2p}
For full reproducibility, this appendix lists the complete deterministic rule sets implemented in \texttt{phoneticspell/g2p.py} across all six languages. In every ordered rule-based frontend, multi-character patterns are matched greedily left-to-right at each index $i$ before contextual and single-character rules.

\subsection{English (\texttt{EnglishG2P} Fallback Rules)}
When a word is absent from \texttt{CMUdict} (\texttt{0.7b}), \texttt{EnglishG2P} strips non-alphabetic characters and applies: (1) silent final \texttt{e} deletion when word length $n > 3$ and the preceding letter is a consonant; (2) ordered multi-character rules (Table~\ref{tab:app_en_g2p}); (3) final \texttt{VCe} lengthening (\texttt{a}$\rightarrow$/e\textsci/, \texttt{e}$\rightarrow$/i/, \texttt{i}$\rightarrow$/a\textsci/, \texttt{o}$\rightarrow$/o\textupsilon/, \texttt{u}$\rightarrow$/u/); (4) contextual palatalization (\texttt{c}$+$\{\texttt{e,i,y}\}$\rightarrow$/s/, \texttt{g}$+$\{\texttt{e,i,y}\}$\rightarrow$/\textdyoghlig/); (5) single-letter defaults (\texttt{a}$\rightarrow$/\ae/, \texttt{b}$\rightarrow$/b/, \texttt{c}$\rightarrow$/k/, \texttt{d}$\rightarrow$/d/, \texttt{e}$\rightarrow$/\textepsilon/, \texttt{f}$\rightarrow$/f/, \texttt{g}$\rightarrow$/\textipa{g}/, \texttt{h}$\rightarrow$/h/, \texttt{i}$\rightarrow$/\textsci/, \texttt{j}$\rightarrow$/\textdyoghlig/, \texttt{k}$\rightarrow$/k/, \texttt{l}$\rightarrow$/l/, \texttt{m}$\rightarrow$/m/, \texttt{n}$\rightarrow$/n/, \texttt{o}$\rightarrow$/\textscripta/, \texttt{p}$\rightarrow$/p/, \texttt{q}$\rightarrow$/k/, \texttt{r}$\rightarrow$/\textturnr/, \texttt{s}$\rightarrow$/s/, \texttt{t}$\rightarrow$/t/, \texttt{u}$\rightarrow$/\textturnv/, \texttt{v}$\rightarrow$/v/, \texttt{w}$\rightarrow$/w/, \texttt{x}$\rightarrow$/ks/, \texttt{y}$\rightarrow$/j/, \texttt{z}$\rightarrow$/z/); and (6) collapsing of consecutive identical letters other than \texttt{e} and \texttt{o}.

\begin{table}[ht]
\centering
\scriptsize
\caption{Ordered multi-character fallback rules in \texttt{EnglishG2P}.}
\label{tab:app_en_g2p}
\begin{tabular}{lp{9.6cm}}
\toprule
Group & Rules (in evaluation order) \\
\midrule
Suffixes \& clusters & \texttt{tion}$\rightarrow$/\textesh\textschwa n/, \texttt{sion}$\rightarrow$/\textyogh\textschwa n/, \texttt{ough}$\rightarrow$/o\textupsilon/, \texttt{augh}$\rightarrow$/\textopeno/, \texttt{eigh}$\rightarrow$/e\textsci/, \texttt{igh}$\rightarrow$/a\textsci/, \texttt{tch}$\rightarrow$/\textteshlig/, \texttt{dge}$\rightarrow$/\textdyoghlig/, \texttt{sch}$\rightarrow$/sk/ \\
Consonant digraphs & \texttt{qu}$\rightarrow$/kw/, \texttt{ph}$\rightarrow$/f/, \texttt{sh}$\rightarrow$/\textesh/, \texttt{ch}$\rightarrow$/\textteshlig/, \texttt{th}$\rightarrow$/\texttheta/, \texttt{wh}$\rightarrow$/w/, \texttt{ng}$\rightarrow$/\textipa{N}/, \texttt{ck}$\rightarrow$/k/, \texttt{dg}$\rightarrow$/\textdyoghlig/, \texttt{gh}$\rightarrow\emptyset$, \texttt{kn}$\rightarrow$/n/, \texttt{gn}$\rightarrow$/n/, \texttt{wr}$\rightarrow$/\textturnr/, \texttt{ps}$\rightarrow$/s/ \\
Vowel digraphs & \texttt{ee}$\rightarrow$/i/, \texttt{ea}$\rightarrow$/i/, \texttt{oo}$\rightarrow$/u/, \texttt{ou}$\rightarrow$/a\textupsilon/, \texttt{ow}$\rightarrow$/o\textupsilon/, \texttt{oi}$\rightarrow$/\textopeno\textsci/, \texttt{oy}$\rightarrow$/\textopeno\textsci/, \texttt{ai}$\rightarrow$/e\textsci/, \texttt{ay}$\rightarrow$/e\textsci/, \texttt{ei}$\rightarrow$/e\textsci/, \texttt{ey}$\rightarrow$/e\textsci/, \texttt{au}$\rightarrow$/\textopeno/, \texttt{aw}$\rightarrow$/\textopeno/ \\
R-colored vowels & \texttt{ar}$\rightarrow$/\textscripta\textturnr/, \texttt{er}$\rightarrow$/\textrhookschwa/, \texttt{ir}$\rightarrow$/\textrevepsilon\textrhoticity/, \texttt{or}$\rightarrow$/\textopeno\textturnr/, \texttt{ur}$\rightarrow$/\textrevepsilon\textrhoticity/ \\
\bottomrule
\end{tabular}
\end{table}

\subsection{Brazilian Portuguese (\texttt{PortugueseG2P})}
Table~\ref{tab:pt_g2p} in Section~\ref{sec:architecture} lists the complete multi-graph, contextual, and non-trivial vowel/consonant rules of \texttt{PortugueseG2P}; the remaining single graphemes map $1$-to-$1$ as \texttt{a}$\rightarrow$/a/, \texttt{b}$\rightarrow$/b/, \texttt{c}$\rightarrow$/k/, \texttt{d}$\rightarrow$/d/, \texttt{e}$\rightarrow$/e/, \texttt{f}$\rightarrow$/f/, \texttt{g}$\rightarrow$/\textipa{g}/, \texttt{i/\'i/y}$\rightarrow$/i/, \texttt{k/q}$\rightarrow$/k/, \texttt{l}$\rightarrow$/l/, \texttt{m}$\rightarrow$/m/, \texttt{n}$\rightarrow$/n/, \texttt{o}$\rightarrow$/o/, \texttt{p}$\rightarrow$/p/, \texttt{r}$\rightarrow$/\textturnr/, \texttt{s}$\rightarrow$/s/, \texttt{t}$\rightarrow$/t/, \texttt{u/\'u}$\rightarrow$/u/, \texttt{v}$\rightarrow$/v/, \texttt{w}$\rightarrow$/w/, and \texttt{z}$\rightarrow$/z/.

\subsection{French (\texttt{FrenchG2P})}
\texttt{FrenchG2P} first strips a single trailing silent consonant (\texttt{s, t, d, x, p, z}) and trailing silent \texttt{e} when word length $n > 3$. It then applies ordered multi-character rules (\texttt{eaux}$\rightarrow$/o/, \texttt{eau}$\rightarrow$/o/, \texttt{tion}$\rightarrow$/sj\textopeno n/, \texttt{\oe u}$\rightarrow$/\o/, \texttt{au}$\rightarrow$/o/, \texttt{ou}$\rightarrow$/u/, \texttt{eu}$\rightarrow$/\o/, \texttt{oi}$\rightarrow$/wa/, \texttt{ai/ei}$\rightarrow$/\textepsilon/, \texttt{ch}$\rightarrow$/\textesh/, \texttt{ph}$\rightarrow$/f/, \texttt{gn}$\rightarrow$/\textipa{J}/, \texttt{qu}$\rightarrow$/k/, \texttt{gu}$\rightarrow$/\textipa{g}/, \texttt{ss}$\rightarrow$/s/), nasal vowel sequences before a consonant or word boundary (\texttt{an/am/en/em}$\rightarrow$/\textscripta/, \texttt{on/om}$\rightarrow$/o/, \texttt{in/im/un}$\rightarrow$/\textepsilon/), contextual rules (\texttt{c/g} before \texttt{e, \'e, \`e, \^e, \"e, i, \^i, \"i, y}$\rightarrow$/s, \textyogh/; intervocalic \texttt{s}$\rightarrow$/z/), single-letter defaults (\texttt{a/\'a/\`a/\"a}$\rightarrow$/a/, \texttt{\^a}$\rightarrow$/\textscripta/, \texttt{e}$\rightarrow$/\textschwa/, \texttt{\'e}$\rightarrow$/e/, \texttt{\`e/\^e/\"e}$\rightarrow$/\textepsilon/, \texttt{h}$\rightarrow\emptyset$, \texttt{i/\^i/\"i/y}$\rightarrow$/i/, \texttt{j}$\rightarrow$/\textyogh/, \texttt{o/\^o}$\rightarrow$/o/, \texttt{\"o/\oe}$\rightarrow$/\o/, \texttt{u/\`u/\^u/\"u}$\rightarrow$/y/, \texttt{\c{c}}$\rightarrow$/s/, \texttt{x}$\rightarrow$/ks/, and standard $1$-to-$1$ consonants), and doubled-consonant collapsing.

\subsection{Turkish (\texttt{TurkishG2P}), Russian (\texttt{RussianG2P}), and Hebrew (\texttt{HebrewG2P})}
\begin{itemize}
    \item \textbf{\texttt{TurkishG2P}:} Applies Turkish case folding (\texttt{\.{I}}$\rightarrow$\texttt{i}, \texttt{I}$\rightarrow$\texttt{\i}) followed by $1$-to-$1$ mapping: \texttt{a/\^a}$\rightarrow$/a/, \texttt{c}$\rightarrow$/\textdyoghlig/, \texttt{\c{c}}$\rightarrow$/\textteshlig/, \texttt{\u{g}}$\rightarrow\emptyset$, \texttt{\i}$\rightarrow$/\textturnm/, \texttt{i/\^i}$\rightarrow$/i/, \texttt{j}$\rightarrow$/\textyogh/, \texttt{\"o}$\rightarrow$/\o/, \texttt{r}$\rightarrow$/\textfishhookr/, \texttt{\c{s}}$\rightarrow$/\textesh/, \texttt{u/\^u}$\rightarrow$/u/, \texttt{\"u}$\rightarrow$/y/, \texttt{y}$\rightarrow$/j/, with remaining Latin consonants mapping $1$-to-$1$.
    \item \textbf{\texttt{RussianG2P}:} Maps word-final genitive \textit{-ogo/-ego} $\rightarrow$ /ovo, jevo/; applies word-final obstruent devoicing (\textit{b, v, g, d, zh, z} $\rightarrow$ /p, f, k, t, \textesh, s/); and maps Cyrillic letters as \textit{a}$\rightarrow$/a/, \textit{b}$\rightarrow$/b/, \textit{v}$\rightarrow$/v/, \textit{g}$\rightarrow$/\textipa{g}/, \textit{d}$\rightarrow$/d/, \textit{e}$\rightarrow$/je/, \textit{yo}$\rightarrow$/jo/, \textit{zh}$\rightarrow$/\textyogh/, \textit{z}$\rightarrow$/z/, \textit{i}$\rightarrow$/i/, \textit{y}$\rightarrow$/j/, \textit{k}$\rightarrow$/k/, \textit{l}$\rightarrow$/l/, \textit{m}$\rightarrow$/m/, \textit{n}$\rightarrow$/n/, \textit{o}$\rightarrow$/o/, \textit{p}$\rightarrow$/p/, \textit{r}$\rightarrow$/\textfishhookr/, \textit{s}$\rightarrow$/s/, \textit{t}$\rightarrow$/t/, \textit{u}$\rightarrow$/u/, \textit{f}$\rightarrow$/f/, \textit{kh}$\rightarrow$/x/, \textit{ts}$\rightarrow$/ts/, \textit{ch}$\rightarrow$/\textteshlig/, \textit{sh/shch}$\rightarrow$/\textesh/, hard/soft signs (\textit{tvjordyj/mjagkij znak}) $\rightarrow\emptyset$, \textit{yery}$\rightarrow$/\textbari/, \textit{\`e}$\rightarrow$/e/, \textit{yu}$\rightarrow$/ju/, \textit{ya}$\rightarrow$/ja/.
    \item \textbf{\texttt{HebrewG2P}:} Maps the $22$ unvocalized consonants and $5$ final (\textit{sofit}) forms $1$-to-$1$: \textit{alef/ayin}$\rightarrow$/\textglotstop/, \textit{bet}$\rightarrow$/b/, \textit{gimel}$\rightarrow$/\textipa{g}/, \textit{dalet}$\rightarrow$/d/, \textit{he}$\rightarrow$/h/, \textit{vav}$\rightarrow$/v/, \textit{zayin}$\rightarrow$/z/, \textit{het/final kaf}$\rightarrow$/x/, \textit{tet/tav}$\rightarrow$/t/, \textit{yod}$\rightarrow$/j/, \textit{kaf/qof}$\rightarrow$/k/, \textit{lamed}$\rightarrow$/l/, \textit{mem/final mem}$\rightarrow$/m/, \textit{nun/final nun}$\rightarrow$/n/, \textit{samekh}$\rightarrow$/s/, \textit{pe}$\rightarrow$/p/, \textit{final pe}$\rightarrow$/f/, \textit{tsadi/final tsadi}$\rightarrow$/ts/, \textit{resh}$\rightarrow$/\textfishhookr/, \textit{shin}$\rightarrow$/\textesh/.
\end{itemize}

\end{document}